\documentclass{article}

\usepackage{graphicx} 
\usepackage{color} 
\usepackage{amsmath}
\usepackage{float}

\numberwithin{equation}{section}

\usepackage[ruled,linesnumbered]{algorithm2e}
\usepackage{appendix}

\usepackage[]{hyperref}
\usepackage[figure]{hypcap}
\usepackage[hyphenbreaks]{breakurl}
\title{\textbf{Mismatch reconstruction theory for unknown measurement matrix in imaging through multimode fiber bending}}
\author{Le Yang}

\makeatletter
\newcommand{\RemoveAlgoNumber}{\renewcommand{\fnum@algocf}{\AlCapSty{\AlCapFnt\algorithmcfname}}}
\newcommand{\RevertAlgoNumber}{\algocf@resetfnum}
\makeatother

\usepackage{cite}

\begin{document}
\maketitle
Multimode fiber imaging requires strict matching between measurement value and measurement matrix to achieve image reconstruction. However, in practical applications, the measurement matrix often cannot be obtained due to unknown system configuration or difficulty in real-time alignment after arbitrary fiber bending, resulting in the failure of traditional reconstruction algorithms. This paper presents a novel mismatch reconstruction theory for solving the problem of image reconstruction when measurement matrix is unknown. We first propose mismatch equation and design matched and calibration solution algorithms to construct a new measurement matrix. In addition, we also provide a detailed proof of these equations and algorithms in the appendix. The experimental results show that under low noise levels, constructed matrix can be used for matched pair in traditional reconstruction algorithms, and reconstruct the original image successfully. Then, we analyze the impact of noise, computational precision and orthogonality on reconstruction performance. The results show that proposed algorithms have a certain degree of robustness. Finally, we discuss the limitations and potential applications of this theory. The code is available: \url{https://github.com/yanglebupt/mismatch-solution}.

\section{Introduction}
Endoscope is a key imaging device that enables direct observation by inserting into the internal cavity of human body and has a wide range of applications in clinical medicine\cite{WOS:000235136400007,WOS:000631182800001,WOS:001511986100004}. Traditional commercial endoscopes typically use single mode fiber bundles as the image transmission medium\cite{WOS:000328078300006,WOS:000481541400016}. The field of view and imaging clarity are limited by the number of fiber and core size, making it difficult to further improve in extremely narrow environments\cite{WOS:000328078300006}. In contrast, a single multimode fiber can not only transmit multiple mode distributions\cite{WOS:000308801100042}, but also has smaller diameter and softer curvature, which show great potential in large field of view and miniaturized endoscopes\cite{WOS:001021361400002,oes:oes-2023-0041}. However, due to the inherent mode dispersion and mode coupling within multimode fiber\cite{Palmieri:24}, the image will appear as an unrecognizable speckle pattern at output end of fiber\cite{WOS:000466458800006}. In order to reconstruct the original image from speckle, researchers have proposed methods including phase conjugation\cite{WOS:000303879700023,WOS:000314806600009}, wavefront shaping\cite{WOS:000369051200017,WOS:000684701400001,WOS:001334679400008}, transmission matrix\cite{WOS:000319339600123,WOS:000366574400003,WOS:000362419900077}, and deep learning\cite{WOS:000449972600108,WOS:000446963000002,WOS:000621765000049,WOS:000749357900016}. But a more complex issue is that the bending of multimode fiber can significantly alter its transmission characteristics, resulting in random changes of speckle pattern. This makes it particularly difficult for reconstruct the original image in dynamic bending multimode fiber scenarios\cite{WOS:000319339600123}. Therefore, achieving high-quality imaging robust to fiber bending has become a key issue for the practical application of multimode fiber endoscopes.
~\\\\
At present, there are mainly two types of multimode fiber imaging methods. One type is transmission matrix-based raster scanning method, which focuses and scans the output light field by measuring the input-output response relationship of fiber\cite{WOS:000275543500006,WOS:000310978500004,WOS:000423776600018}. Another type is speckle imaging-based compressed sensing method, which records a series of speckle patterns, followed by their projection and reconstruction during imaging stage\cite{WOS:000448939000065,WOS:000466160900091,WOS:000489024000008,WOS:000471824700002}. However, these methods are highly sensitive to bending and disturbances. Once the shape of fiber changes, original calibration becomes invalid, resulting in a sharp decline in the quality of reconstructed image or even the inability to reconstruct. To overcome these problems, mathematical model can be established under certain specific fibers to describe the propagation of light after fiber bending and predict its transmission matrix\cite{WOS:000358737200016}. 
However, mathematical model requires precise knowledge of fiber layout, and the model cannot be generalized to any bending under any fiber. Feedback compensation technology calibrates transmission matrix in real time and refocuses light by obtaining signal from the distal end of fiber\cite{WOS:000319339600123,WOS:000325547200074,WOS:000366574400003,WOS:000387537600022}, which requires additional feedback equipment and system. The statistical averaging method smooths fluctuation of local disturbances by sampling and averaging the transmission matrices\cite{WOS:000310978500004,WOS:000489024000008,WOS:000530854700091}, but this cannot overcome disturbances beyond local correlations. Optical memory effect also can be used to search invariant in fiber bending, such as speckle autocorrelation\cite{WOS:000384715800062} and isoplanatic patch\cite{WOS:000665038900019}, to achieve image reconstruction. However, the imaging capability of this method is limited by memory effect range.
~\\\\
Deep learning, as an end-to-end neural network model, can reconstruct original image from speckle pattern or measurement value without prior knowledge of fiber transmission characteristics, which has been widely used in multimode fiber and its bending imaging. The most common method is to collect speckle-image pairs on multiple fiber bending configurations for mixed training, so that the trained models can adapt to unknown bending configurations\cite{WOS:000476652500013,WOS:000650530200001,WOS:000680277500001}. By using a variational autoencoder with gaussian mixture latent space, speckles with different bending perturbations are projected onto the same clusters in the latent space, and then learn the reconstruction mapping between cluster and original image\cite{WOS:001030642400042}. In addition, some physical prior knowledge can also be incorporated. Due to the high similarity of speckle autocorrelation within memory effect range\cite{WOS:000384715800062}, and the redundancy of speckle, that is, the autocorrelation maps generated by different subregions of same speckle are similar\cite{WOS:000343145200012}, using only a single bending configuration to train UNet and reconstruct object images from speckle autocorrelation maps can achieve good resistance to fiber bending\cite{WOS:000904474500006}. However, the training cost of these methods is high, and it is still difficult to reconstruct when the target images or bends beyond the training distribution.
~\\\\
In this paper, in order to better solve the difficulty of image reconstruction caused by the unknown measurement matrix due to arbitrary bending of multimode fiber, we first solve the equation systems of known measurement and unknown measurement under same image, and obtain the alternative solution of unknown measurement matrix as the mismatch equation. On this basis, a matched solution algorithm is proposed by combining pre-measure and error iteration method. Then use multiplier property of mismatch equation to improve the matched solution algorithm. Afterwards, based on orthogonal basis images, we extend the mismatch equation to the calibration equation and propose a calibration solution algorithm. The solvability of these equations and the convergence of these algorithms are proofed in the appendix. Then, we conduct simulation experiments on these two types of algorithms, including convergence analysis, matched solution reconstruction, calibration solution reconstruction and impact of computational precision. The results show that using the constructed measurement matrix by these algorithms in image reconstruction algorithm can not only successfully reconstruct the original image, but also has certain robustness to the noise and orthogonality. Compared with previous work, this theory does not require precise fiber mathematical modeling, complex feedback mechanism or high training cost. It only needs to use the same additional measurement steps and combine proposed algorithms to construct a new measurement matrix after fiber bending to reconstruct the original image. And it is applicable to any image and bending. This theory establishes the theoretical foundation and algorithmic tools for portable and stable dynamic imaging of multimode fiber.

\newpage

\section{Methods}
\subsection{Matched Solution of Unknown Measurement Matrix}
In general, regardless of how the multimode fiber is bent, the measurement matrix $A_u$ and the measurement value $y$ are always a matched pair that satisfies the imaging equation $y=A_u x+\epsilon_1$. However, different bending configurations correspond to different measurement matrices, which are difficult to measure in real-time in actual imaging environments. And existing compressed sensing algorithms $G$ cannot reconstruct the original image without the measurement matrix. Therefore, the goal of this study is to find a constructible new measurement matrix $A_{recv}$ making ($y$, $A_{recv}$) is a matched pair, and then the original image can be reconstructed using the existing reconstruction algorithm $x^{*}=G(y, A_{recv})$. It is also equivalent to the optimization problem as follows:

\begin{equation}
\label{eq:goal}
A_{recv} = \arg\min_{\Lambda} ||y- \Lambda x||_2^2
\end{equation}

To solve this optimization problem, assuming that there exists a known measurement $y_0=Ax+\epsilon_2$ for the same image $x$. Then the optimization problem is transformed into the solution of matrix equations as follows:

\begin{equation}
\label{eq:mgoal}
\begin{cases}
y=A_u x+\epsilon_1\\
y_0=Ax+\epsilon_2
\end{cases}
\Rightarrow A_{recv} = \varphi(y,y_0,A) \sim A_u
\end{equation}

Due to the solution $\varphi$ depend on a measurement matrix $A$ that does not match the target measurement value $y$, this study refers to the solution of such problems as mismatch reconstruction theory, which is also known as mismatch imaging when applied to multimode fiber imaging. According to the derivation in \hyperref[appe:A]{Appendix.\ref{appe:A}}, a general solution of $\varphi$ as shown in \hyperref[eq:Arecv]{Eq.\ref{eq:Arecv}} when without nosie ($\epsilon_1=0, \epsilon_2=0$). $A_{recv}^{(y_0, y)}$ is a symbol notation for that general solution. When $y_0$ is a constant, it can be abbreviated as $A_{recv}^{y}$. Meanwhile, by substituting $A_{recv}^{(y_0, y)}x \equiv y$, the correctness of the general solution can be quickly verified.

\begin{equation}
\label{eq:Arecv}
\varphi(y,y_0,A)=A_{recv}^{(y_0, y)} = \frac{1}{y_0^T\Sigma y_0}yy_0^T\Sigma A (\Sigma \ne 0)
\end{equation}

This general solution is the most important equation in all subsequent algorithms and even the entire mismatch reconstruction theory. To emphasize this, this study refers to this general solution as \textbf{Mismatch Equation}. It means relationship between measurement value and measurement matrix is not one-to-one when $x$ is fixed, indicating that there is not only one solution for \hyperref[eq:goal]{Eq.\ref{eq:goal}}. And the general solution of mismatch equation $A_{recv}^{(y_0, y)}$ is not equal to the real $A_u$. However, mismatch equation cannot be practically applied limited by the two conditions that $(y_0, A)$ is a matched pair under the same image $x$ and without noise. Therefore, \textbf{Error Iteration Algorithm} as described in \hyperref[algo:1]{Algorithm.\ref{algo:1}} is proposed to construct $A_{recv}$ by breaking above limits.

\newpage

\begin{algorithm}[!h]
	\caption{$A_{recv}$ for $y$}
	\label{algo:1}
	\LinesNumbered
	\KwIn{$y$}
	\KwOut{$A_{recv}$}
        \textbf{Parameters:}$\; epoch, \; A, \; PM_{image}$ \\
        $y_0=A * PM_{image}$ \\
        $A_{recv}=0$ \\
        $\Sigma=(AA^T)^{-1}$ \\
        $e_y=y$ \\ 
	\For{$i\leftarrow 1 \; to \; epoch$}{
          $A_{recv}^{e_y} = \frac{1}{y_0^T\Sigma y_0}\;e_y\;y_0^T\Sigma A$ \\ 
	       $A_{recv} += A_{recv}^{e_y}$ \\
          $e_y$ = $y$ - $speckle\_measure$($A_{recv} \rightarrow x$)
	}
\end{algorithm}

The algorithm first performs $y_0=A * PM_{image}$, which named as \textbf{Pre-Measure} and is completely known and computable. $PM_{image}$ is a pre-measure image that can be any non-zero image, so in this case $(y_0, A)$ is a matched pair under the pre-measure image $PM_{image}$ instead of the measured image $x$. Then, initialize $A_{recv}$ to 0, specific solution of mismatch equation to $\Sigma=(AA^T)^{-1}$ (Derivation is provided in \hyperref[appe:A]{Appendix.\ref{appe:A}}) and error $e_y$ to measurement value $y$. In each iteration, calculate mismatch equation $A_{recv}^{e_y}$ by current error $e_y$ and accumulate it onto $A_{recv}$. Then use accumulated $A_{recv}$ to measure image $x$ obtaining the current measurement value. Finally update error $e_{y}$ as the result of target measurement value $y$ minus current measurement value. Convergence proof of this algorithm with measurement noise is provided in \hyperref[appe:B]{Appendix.\ref{appe:B}}. However, it still has two disadvantages as follows:

\begin{itemize}
\label{eq:disadv}
\item Each iteration requires measurement once, which introduces many additional measurements.
\item Need to run the algorithm again to obtain a new $A_{recv}$ for each different measured image, even if the fiber bending configuration remains unchanged.
\end{itemize}

To reduce additional measurements, need to use \textbf{Multiplier Property} of mismatch equation as shown in \hyperref[eq:m]{Eq.\ref{eq:m}}.

\begin{equation}
\label{eq:m}
\begin{aligned}
A_{recv}^{e_y}x = \frac{1}{y_0^T\Sigma y_0}&e_yy_0^T\Sigma A x = k(x)*e_y \\
k(x)=&\frac{y_0^T\Sigma A x}{y_0^T\Sigma y_0} \in R
\end{aligned}
\end{equation}

This means that when using $A_{recv}^{e_y}$ measure different images $x$, the results are proportional, and the coefficient $k$ is only related to pre-measure and measured image not to iterative error $e_y$. Therefore, coefficient $k$ is a constant throughout the entire iteration. And since $k(PM_{image})=1$, use $A_{recv}^{y_0}$ to multiply $x$ and $PM_{image}$ separately can solve the coefficient $k(x)$ as follows:

\begin{equation}
\label{eq:sk}
\begin{aligned}
y'=&A_{recv}^{y_0}x=k(x)y_0 \\
y_{pm}=A_{recv}^{y_0}&PM_{image}=k(PM_{image})y_0
\end{aligned}
\Rightarrow k(x) = \frac{y'}{y_{pm}}
\end{equation}

Therefore, only one initial measurement is needed to calculate the coefficient $k$. In each iteration, replace measurement with $k*A_{recv}*PM_{image}$. This result in an improved error iteration algorithm as described in \hyperref[algo:2]{Algorithm.\ref{algo:2}}, which is \textbf{\textit{Matched Solution of Unknown Measurement Matrix}}.

\begin{algorithm}[!h]
	\caption{Matched Solution of Unknown Measurement Matrix}
	\label{algo:2}
	\LinesNumbered
	\KwIn{$y$}
        \KwOut{$A_{recv}$}
        \textbf{Parameters:}$\; epoch, \; A, \; PM_{image}$ \\
	$(y_0, \; A_{recv}^{y_{0}})=Initialize(epoch, \; A, \; PM_{image})$ \\
        $A_{recv}=A_{recv}^{y_{0}}$ \\
        $\Sigma=(AA^T)^{-1}$ \\
        $y' = speckle\_measure(A_{recv} \rightarrow x)$ \\
        $y_{pm} = A_{recv} * PM_{image}$ \\
        $k=y' / y_{pm} $ \\
        $e_y=y - y'$ \\
        \For{$i\leftarrow 1 \; to \; epoch$}{
            $A_{recv}+=\frac{1}{y_0^T\Sigma y_0}\;e_y\;y_0^T\Sigma A$ \\
            $e_y$ = $y$ - $k*A_{recv}*PM_{image}$
        }
\end{algorithm}
\begin{algorithm}[!h]
        \renewcommand{\thealgocf}{2.1}
	\caption{Initialize $A_{recv}^{y_0}$}
	\LinesNumbered
        \label{algo:pm}
	\KwIn{$epoch, \; A, \; PM_{image}$}
	\KwOut{$y_0, \; A_{recv}^{y_0}$}
        $y_0=A * PM_{image}$ \\
        $A_{recv}^{y_0}=0$ \\
        $\Sigma=(AA^T)^{-1}$ \\
        $e_y=y_0$ \\ 
	\For{$i\leftarrow 1 \; to \; epoch$}{	
	       $A_{recv}^{y_0} += \frac{1}{y_0^T\Sigma y_0}\;e_y\;y_0^T\Sigma A$ \\
          $e_y$ = $y_0$ - $A_{recv}^{y_0} * PM_{image}$
	}
\end{algorithm}

The algorithm first generate $(y_0, A_{recv}^{y_0})$ by $Initialize$ as described in \hyperref[algo:pm]{Algorithm.\ref{algo:pm}}. Theoretically $A_{recv}^{y_0}$ can be directly calculated by mismatch equation, but using float32 to calculate the matrix will introduce precision error unavoidably. So here still use error iteration algorithm to offset the calculation precision error and construct $A_{recv}^{y_0}$. Next, initialize $A_{recv}$ to $A_{recv}^{y_0}$ and specific solution of mismatch equation to $\Sigma=(AA^T)^{-1}$, then calculate the coefficient $k$ by \hyperref[eq:sk]{Eq.\ref{eq:sk}} and initialize error $e_y$ to $y-y'$. In each iteration, it is purely computation without the need for measurement. When existing measurement noise, as shown in \hyperref[eq:ek]{Eq.\ref{eq:ek}}, the coefficient $k$ is a valid approximation of $k(x)$.

\begin{equation}
\label{eq:ek}
\begin{aligned}
y'=speckle\_measure(A_{recv}^{y_0} \rightarrow x)=k(x)y_0+\epsilon \\
k=\frac{y'}{y_{pm}}=\frac{k(x)y_0+\epsilon}{k(PM_{image})y_0}\approx k(x)
\end{aligned}
\end{equation}

After obtaining the coefficient approximation, there is no need to measure image in each iteration, just replace simply as follows:

\begin{equation}
\label{eq:rk}
\begin{aligned}
k*A_{recv}^{e_y}*PM_{image} &= \frac{k(x)y_0+\epsilon}{k(PM_{image})y_0} * k(PM_{image})e^y \\
&=k(x)e^y+\frac{e^y}{y_0}\epsilon \\
&=k(x)e^y+\epsilon' \\
&\approx speckle\_measure(A_{recv}^{e_y} \rightarrow x)
\end{aligned}
\end{equation}

This means that after the replacement, the measurement noise is changed from $\epsilon$ to $\epsilon'=\frac{e^y}{y_0}\epsilon$. Due to $e^y$ converges to 0, $\epsilon'$ should also converge to 0. So \hyperref[algo:2]{Algorithm.\ref{algo:2}} still can be converged.

\subsection{Calibration Solution of Unknown Measurement Matrix}
\label{sec:b}

Consider the \hyperref[eq:disadv]{second disadvantage} left over from the previous section. The main reason of this disadvantage is that constructed $A_{recv}$ by error iteration algorithm is not equal to $A_u$, it is only a matched solution of target measurement value that depends on measured image and fiber bending configuration. For each different measured image, even if the fiber bending configuration remains unchanged, it will result in different measurement values, and need to rerun the algorithm again to construct a new $A_{recv}$. Therefore, this section attempts to further provide a calibration method so that one calibration can be applied to any measured image under the same fiber bending configuration. This type of method is named \textbf{\textit{Calibration Solution of Unknown Measurement Matrix}}.
~\\\\
Firstly, assuming measured image located in $D$-dimension space and can be linearly represented by orthogonal basis images $\{\mathbf{x_{i}}\}_{i=1}^{D}$ in this space.
\begin{equation}
\label{eq:orthogonalbase}
x=\sum_{i=1}^{D} b_i\bf{x_i}
\end{equation}

Conduct pre-measure and unknown measure on all basis images.
\begin{equation}
\label{eq:pre-unknown}
\begin{cases}
y_{i}^0=A\mathbf{x_i}\\
y_{i}=A_u\mathbf{x_i}+\epsilon_{i}
\end{cases}
\end{equation}

According to \hyperref[appe:C]{Appendix.\ref{appe:C}}, the solution of this system of equations is:

\begin{equation}
\label{eq:es}
A_{recv}=\sum_{j}^{D} A_{recv}^{(y_{j}^0,y_j)}
\end{equation}

To emphasize this, this study refers to this solution as \textbf{Calibration Equation}. $A_{recv}^{(y_{j}^0,y_j)}$ is the mismatch equation, and its special solution $\Sigma$ must satisfy following condition:

\begin{equation}
\label{eq:condition}
Y\Sigma Y^T=E
\end{equation}

$E$ is the identity matrix, $Y\in R^{D \times M}$ consists of the pre-measurement value $\{(y_j^0)^T\in R^{1 \times M}\}_{j=1}^{D}$ of all basis images in rows. Generally speaking, the dimension $D$ of image space is much larger than the number of measurements $M$, resulting in $Y$ is a column full rank matrix. Therefore, it is easy to find a solution $\Sigma$ by pseudo-inverse as follows:
\begin{gather}
\label{eq:consol}
\begin{aligned}
\Sigma &= Y_{left}^{\dagger}(Y^T)_{right}^{\dagger} \\
&=(Y^TY)^{-1}Y^TY(Y^TY)^{-1} \\
&=(Y^TY)^{-1}
\end{aligned}
\end{gather}

$\dagger$ is the marker of pseudo-inverse. And, a new calibration algorithm described in \hyperref[algo:calidate-mspace]{Algorithm.\ref{algo:calidate-mspace}} is proposed to construct $A_{recv}$ based on calibration equation.

\RevertAlgoNumber
\begin{algorithm}[!h]
        \renewcommand{\thealgocf}{3}
	\caption{Calibration Solution of Unknown Measurement Matrix}
	\label{algo:calidate-mspace}
	\LinesNumbered
	\KwOut{$A_{recv}$}
        \textbf{Parameters:} $A$ \\
        $A_{recv}=0$ \\
        $\Sigma=(Y^TY)^{-1}$ \\
        $Q=qr\_decomposition(A^T)$ \\ 
        $Y=AQ$ \\
        $Y_u=speckle\_measure(A_u, Q)$ \\
	\For{$i\leftarrow 1 \; to \; M$}{	
          $y_0 = Y(:,i)$\\
          $y=Y_u(:,i)$\\
	   $A_{recv} += \frac{1}{y_0^T\Sigma y_0}\;y\;y_0^T\Sigma A$ \\
	}
\end{algorithm}

In order to obtain orthogonal basis images, conduct QR decomposition on the transpose of pre-measurement matrix $A^T$, and $M$ column vectors of the $Q\in R^{N\times M}$ matrix are taken as the basis images. Note that the selection of these basis images is only for the purpose of quickly verify the correctness of calibration algorithm. And it is not a complete basis set and cannot represent all the images to be constructed. In practical applications, should further explore the selection of basis images.

\section{Experiments And Results}
\begin{figure}[htbp]
  \centering
  \setlength{\abovecaptionskip}{-0.15cm}
  \includegraphics[scale = 0.35]{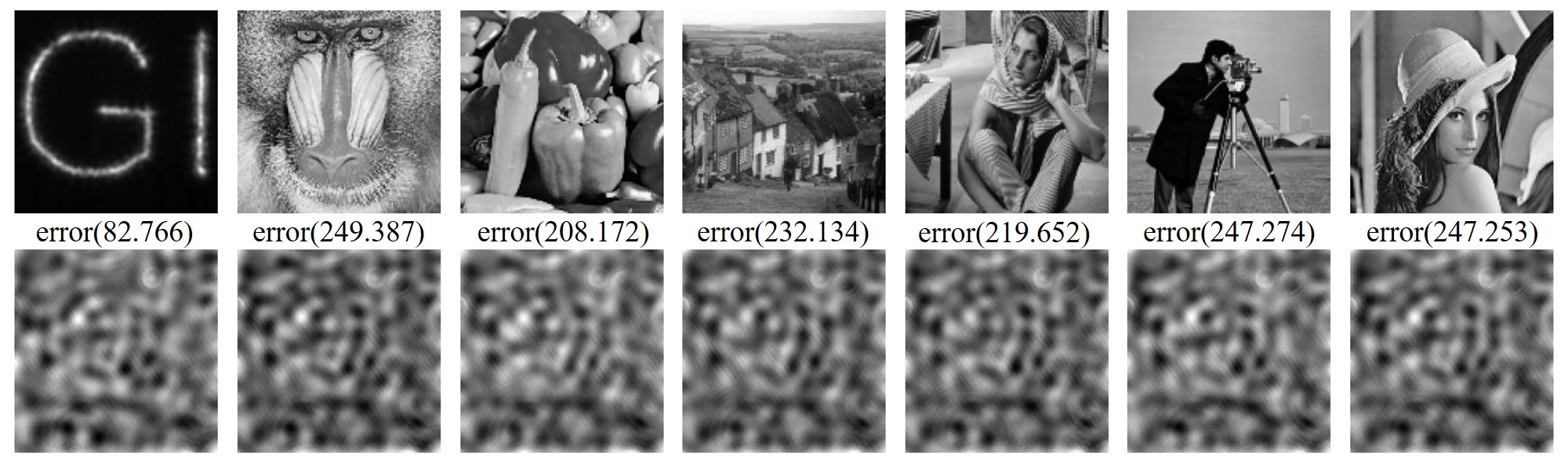}
  \caption{Results of mismatch reconstruction.}
  \label{fig:mismatch}
\end{figure}

This study simulate bending by vertically offset the middle part of multimode fiber by 25 units, and measure images of $N(128\times 128)$ pixels $M(2500)$ times. Collect fiber output end speckle patterns to form unknown measurement matrix $A_u$. The speckle patterns collected from a straight fiber are used to form the pre-measurement matrix $A$. Note that the pre-measurement matrix can be replaced by random gaussian distribution matrix. The reconstruction algorithm is gradient projection sparse reconstruction\cite{WOS:000265494900006}. Algorithms are implemented on RTX2080 Ti (11GB), Python, and use Tensor for matrix operations.
~\\\\
Use the unknown measurement matrix $A_u$ measure images to obtain the measurement value $y$ as the reconstruction target. Firstly, conduct mismatch reconstruction as a comparison benchmark, using the pre-measurement matrix $A$ as the constructed new measurement matrix $A_{recv}$ for reconstruction $G(y, A_{recv}=A)$. Define the error as shown in \hyperref[eq:ee]{Eq.\ref{eq:ee}} to measure matching degree of constructed measurement matrix by algorithms. The larger error, the lower matching degree. The results of mismatch reconstruction are shown in \hyperref[fig:mismatch]{Fig.\ref{fig:mismatch}}. The original images cannot be reconstructed and the error is also large.
\begin{equation}
\label{eq:ee}
error = E||y-A_{recv}x||
\end{equation}

\subsection{Convergence Analysis}
\begin{figure}[H]
  \centering
  \setlength{\abovecaptionskip}{-0.05cm}
  \includegraphics[scale = 0.625]{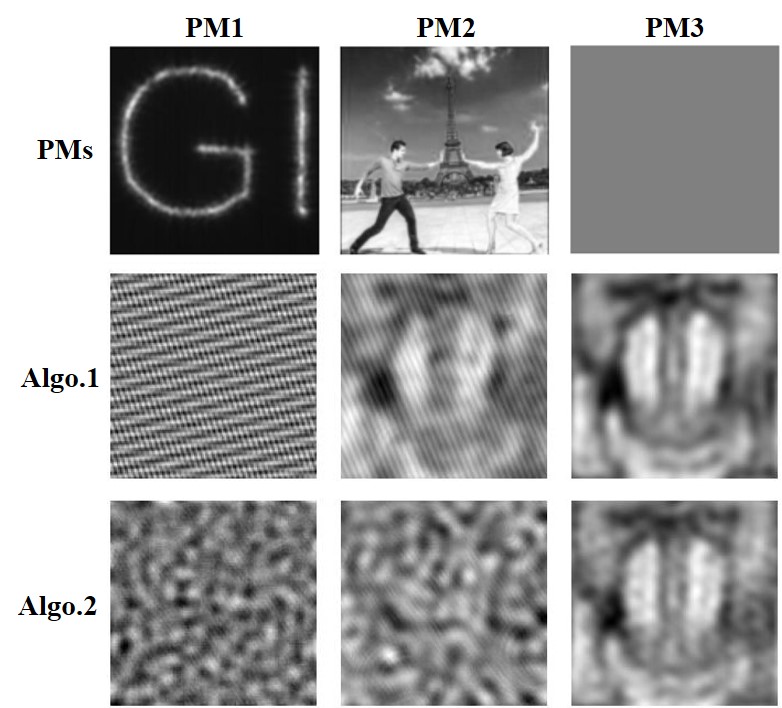}
  \caption{Reconstruction results of Baboon image using constructed $A_{recv}$ by Algo.1 and Algo.2 on three different $PM_{image}$.}
  \label{fig:recv-res}
\end{figure}

Three different $PM_{image}$(PM1, PM2, PM3) are used for both error iteration algorithms: \hyperref[algo:1]{Algorithm.\ref{algo:1}}(Algo.1) and \hyperref[algo:2]{Algorithm.\ref{algo:2}}(Algo.2) to construct $A_{recv}$. And the reconstruction results of Baboon image are shown in \hyperref[fig:recv-res]{Fig.\ref{fig:recv-res}}. When use the pure gray PM3, relatively clear original image is reconstructed both in Algo.1 and Algo.2. When use PM2, original image is reconstructed only in Algo.1, and the clarity decreased. When use PM1, original image cannot be reconstructed both in Algo.1 and Algo.2. 

\begin{figure}[htbp]
  \centering
  \setlength{\abovecaptionskip}{-0.05cm}
  \includegraphics[scale = 0.4]{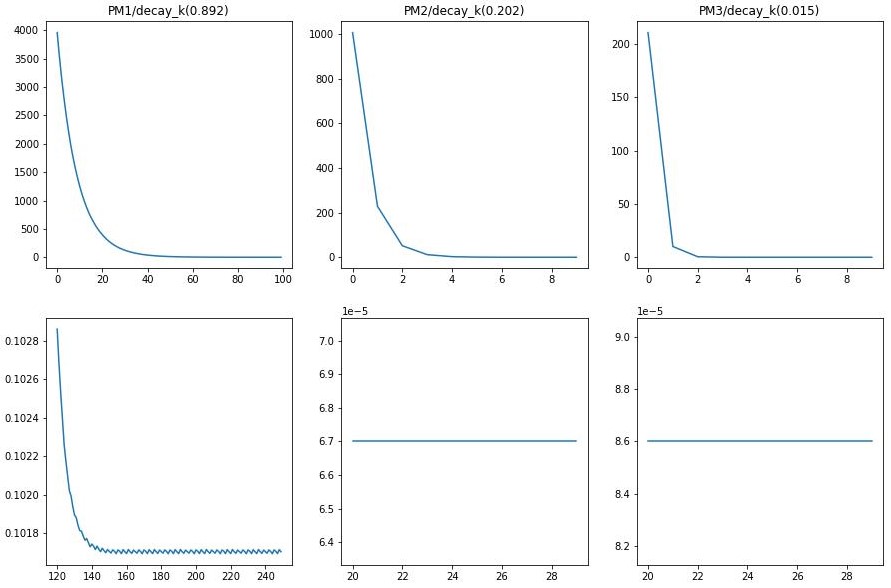}
\caption{Error curves of Algo.1. The horizontal axis is the number of iteration and the vertical axis is error value. Same column shows the error curve in different iteration intervals.}
  \label{fig:algo1-curve}
\end{figure}

\begin{figure}[H]
  \centering
  \setlength{\abovecaptionskip}{-0.05cm}
  \includegraphics[scale = 0.4]{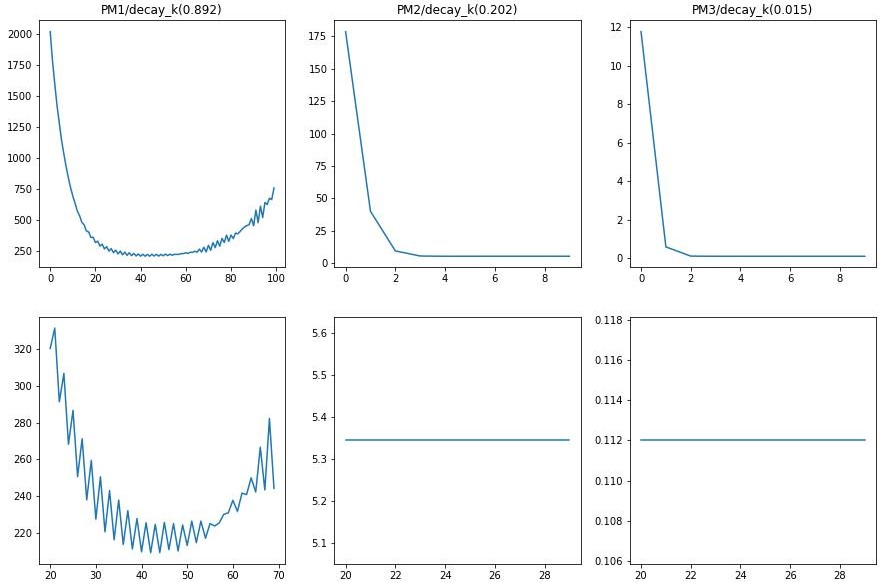}
\caption{Error curves of Algo.2. The horizontal axis is the number of iteration and the vertical axis is error value. Same column shows the error curve in different iteration intervals.}
  \label{fig:algo2-curve}
\end{figure}

According to the derivation in \hyperref[appe:B]{Appendix.\ref{appe:B}}, the convergence factor $||k_{\varepsilon}||$(decay\_k) of this type of algorithm is shown in \hyperref[eq:ske]{Eq.\ref{eq:ske}}. The closer the convergence factor approaches 0, the better the convergence effect. The error curves during iteration process of Algo.1 and Algo.2 are shown in \hyperref[fig:algo1-curve]{Fig.\ref{fig:algo1-curve}} and \hyperref[fig:algo2-curve]{Fig.\ref{fig:algo2-curve}} respectively. 

\begin{equation}
\label{eq:ske}
||k_{\varepsilon}||=||1-\frac{y_0^T\Sigma Ax}{y_0^T\Sigma y_0}||
\end{equation}

As shown in \hyperref[fig:algo1-curve]{Fig.\ref{fig:algo1-curve}}, when Algo.1 uses PM2 and PM3, the error converges to the level of 1e-5 after 20 iterations, and so the original image can be reconstructed naturally. However, when Algo.1 uses PM1, the error only converges to around 0.1 after 140 iterations, and so the original image cannot be reconstructed. Similarly, as shown in \hyperref[fig:algo2-curve]{Fig.\ref{fig:algo2-curve}}, when Algo.2 uses PM3, the error converges to around 0.1 after 20 iterations, and so the reconstructed image is not clear enough. However, when Algo.2 uses PM2, the error converges to around 5.35 after 20 iterations, and so the original image cannot be reconstructed. And when Algo.2 uses PM1, the error first decrease and then increase, which is not converged and so cannot reconstruct the original image naturally. So, Algo.1 and Algo.2 have different convergence errors for the reconstructed image. On the other hand, the convergence effect on different PMs is ranked as PM3$>$PM2$>$PM1 in both Algo.1 and Algo.2, which is consistent with the ranking of convergence factor 0.015$<$0.202$<$0.892.

\newpage

\subsection{Matched Solution Reconstruction}
\begin{figure}[H]
  \centering
  \setlength{\abovecaptionskip}{-0.05cm}
  \includegraphics[scale = 0.23]{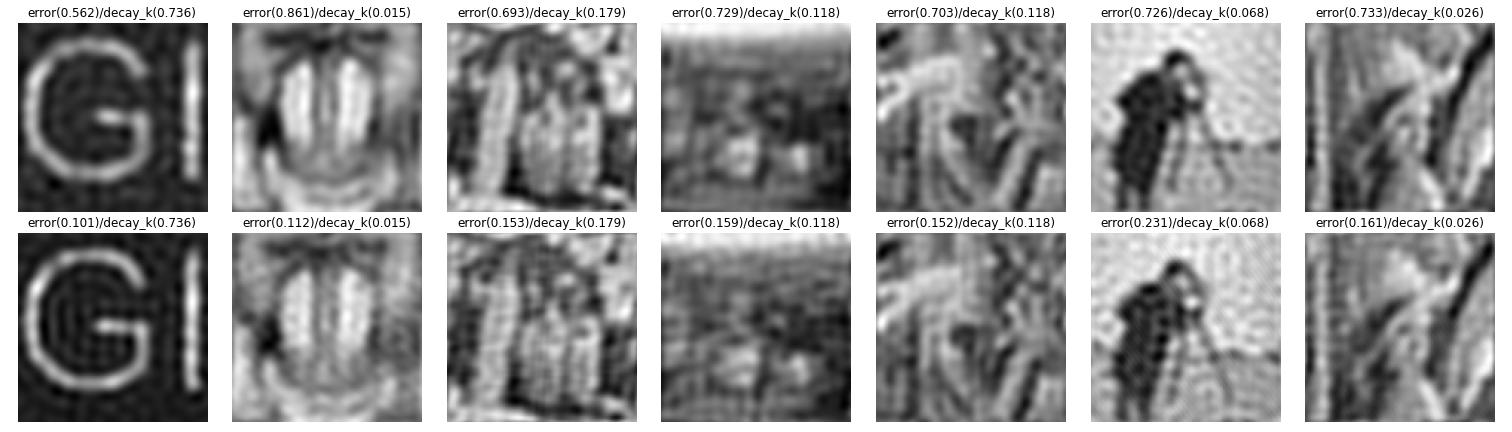}
\caption{Reconstruction results of seven images using constructed $A_{recv}$ by Algo.1 (first line and its error level is 1e-4) and Algo.2 (second line) on PM3.}
  \label{fig:recv-res-row1}
\end{figure}

Reconstruct seven images using constructed $A_{recv}$ by Algo.1 and Algo.2 on PM3, and the results are shown in \hyperref[fig:recv-res-row1]{Fig.\ref{fig:recv-res-row1}}. It can be seen that the original images are successfully reconstructed, and the convergence factors are less than 0.18 except for the GI image, which is 0.736, showing good convergence effects.

Add gaussian noise $\epsilon$ of \hyperref[eq:noise]{Eq.\ref{eq:noise}} to the $speckle\_measure$ function in the simulation experiment, where $\sigma$ is the standard deviation.

\begin{equation}
\label{eq:noise}
\epsilon = \sigma * N(0,1)
\end{equation}

\begin{figure}[H]
  \vspace{-0.5pt}
  \centering
  \setlength{\abovecaptionskip}{-0.05cm}
  \includegraphics[scale = 0.6]{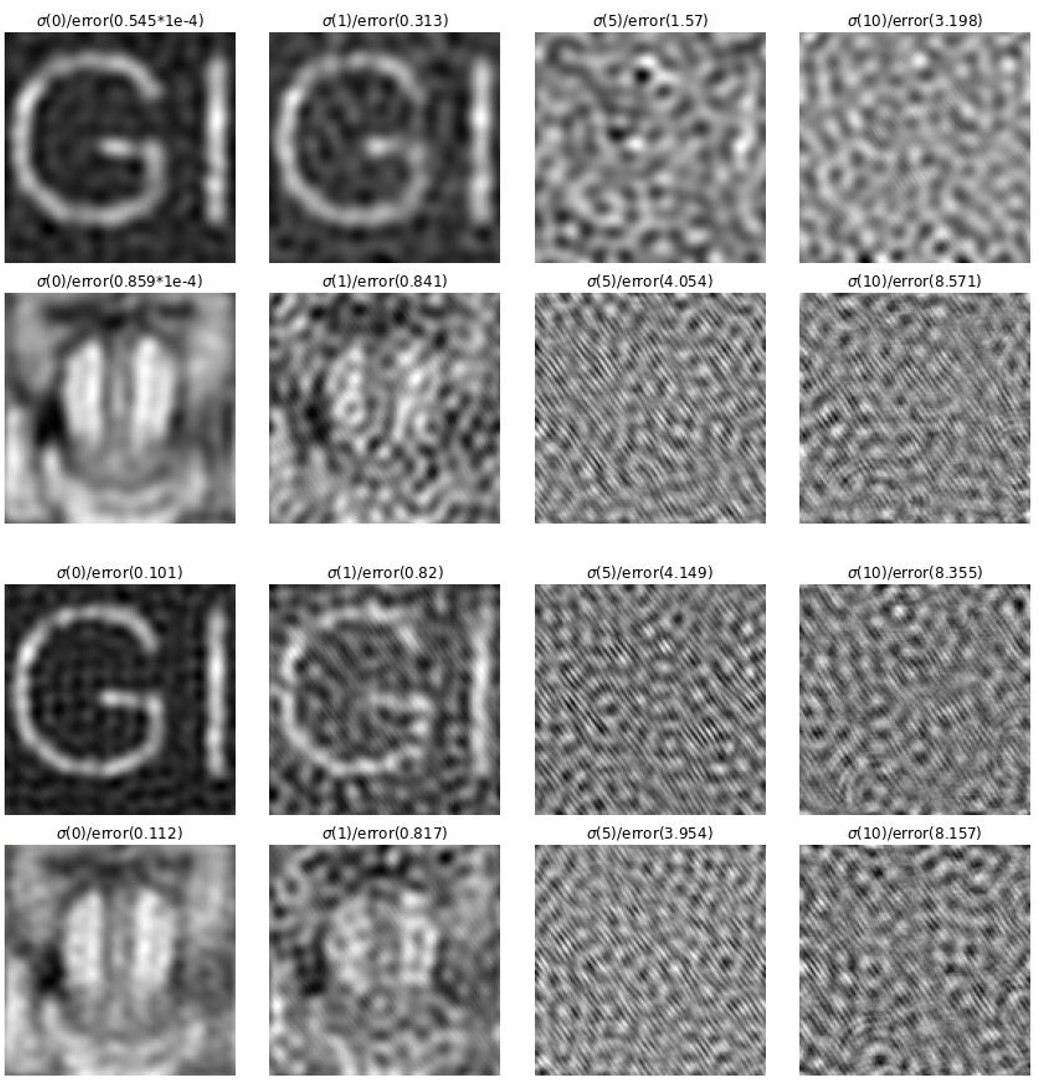}
\caption{Reconstructed GI and Baboon images using constructed $A_{recv}$ by Algo.1 (1-2 lines) and Algo.2 (3-4 lines) on PM3.}
  \label{fig:noise}
\end{figure}

\begin{figure}[H]
  \centering
  \setlength{\abovecaptionskip}{-0.05cm}
  \includegraphics[scale = 0.265]{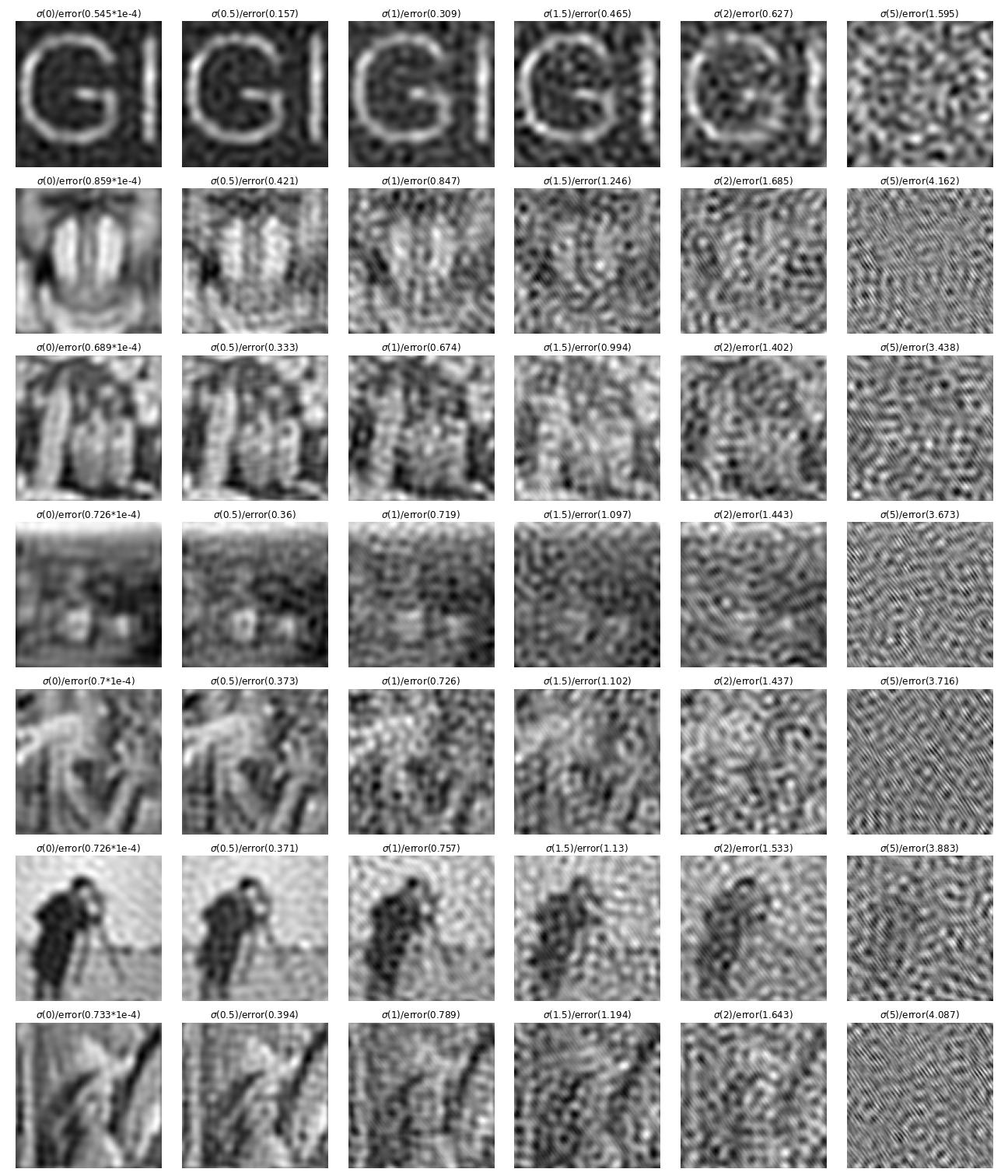}
\caption{Reconstruction results of seven images using constructed $A_{recv}$ by Algo.1 on PM3 and different gaussian noise levels.}
  \label{fig:recv-res-algo1}
\end{figure}

\begin{figure}[H]
  \centering
  \setlength{\abovecaptionskip}{-0.05cm}
  \includegraphics[scale = 0.265]{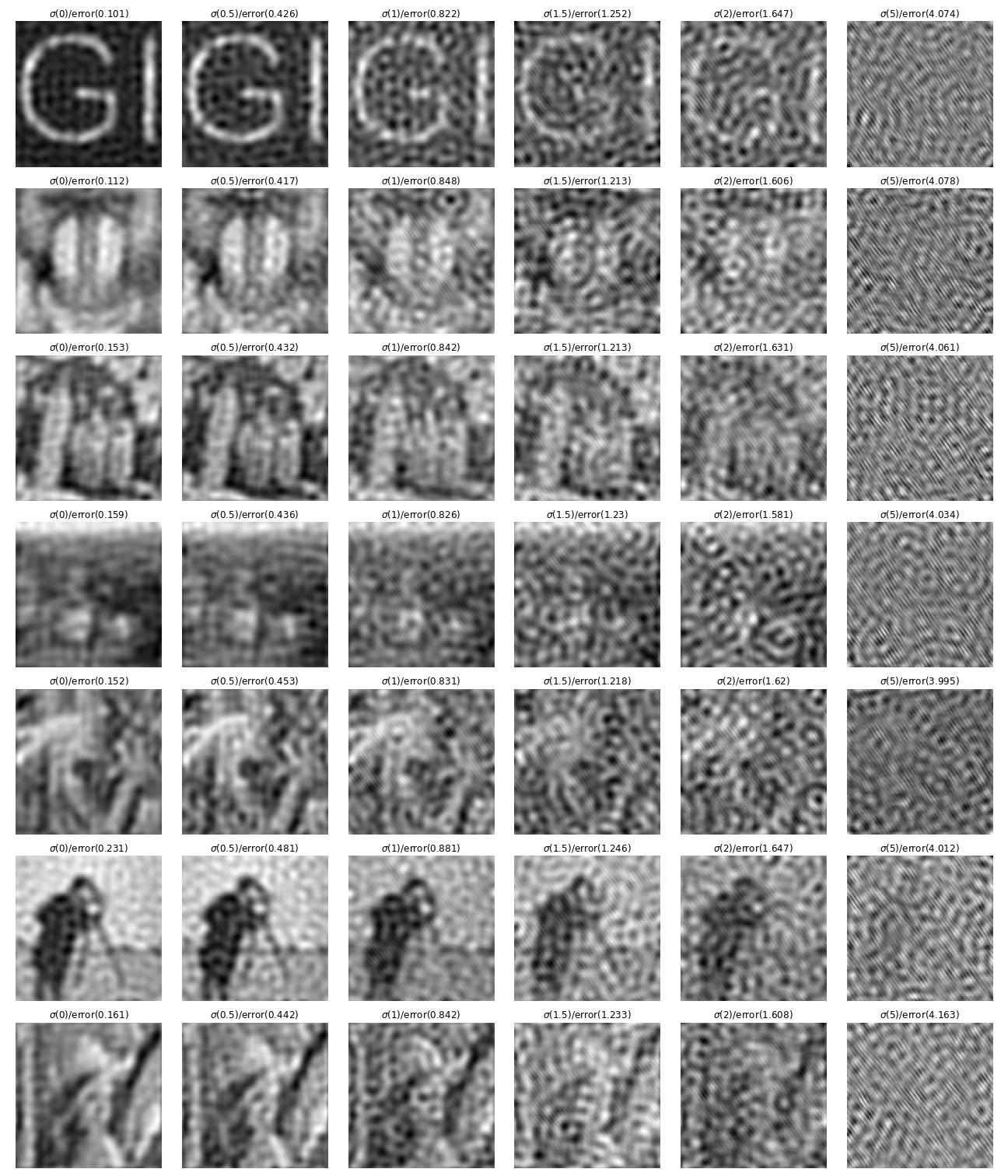}
\caption{Reconstruction results of seven images using constructed $A_{recv}$ by Algo.2 on PM3 and different gaussian noise levels.}
  \label{fig:recv-res-algo2}
\end{figure}

Partial reconstruction results are shown in \hyperref[fig:noise]{Fig.\ref{fig:noise}}. As the standard deviation increases, the convergence error also continues to increase. When $\sigma\leq 1$, the original images still can be reconstructed, but when $\sigma\geq 5$, the original images cannot be reconstructed. More detailed noise level $\sigma$ sequence includes 0, 0.5, 1, 1.5, 2, 5 and its results are shown in \hyperref[fig:recv-res-algo1]{Fig.\ref{fig:recv-res-algo1}} and \hyperref[fig:recv-res-algo2]{Fig.\ref{fig:recv-res-algo2}}, which as additional supplementary in \hyperref[appe:D]{Appendix.\ref{appe:D}}. It can be seen that $\sigma=2$ is already the maximum noise level that can reconstruct original images, indicating that the anti-noise ability of constructed $A_{recv}$ by such algorithms is weak.

\subsection{Calibration Solution Reconstruction}
\begin{figure}[H]
  \centering
  \setlength{\abovecaptionskip}{-0.05cm}
  \includegraphics[scale = 0.225]{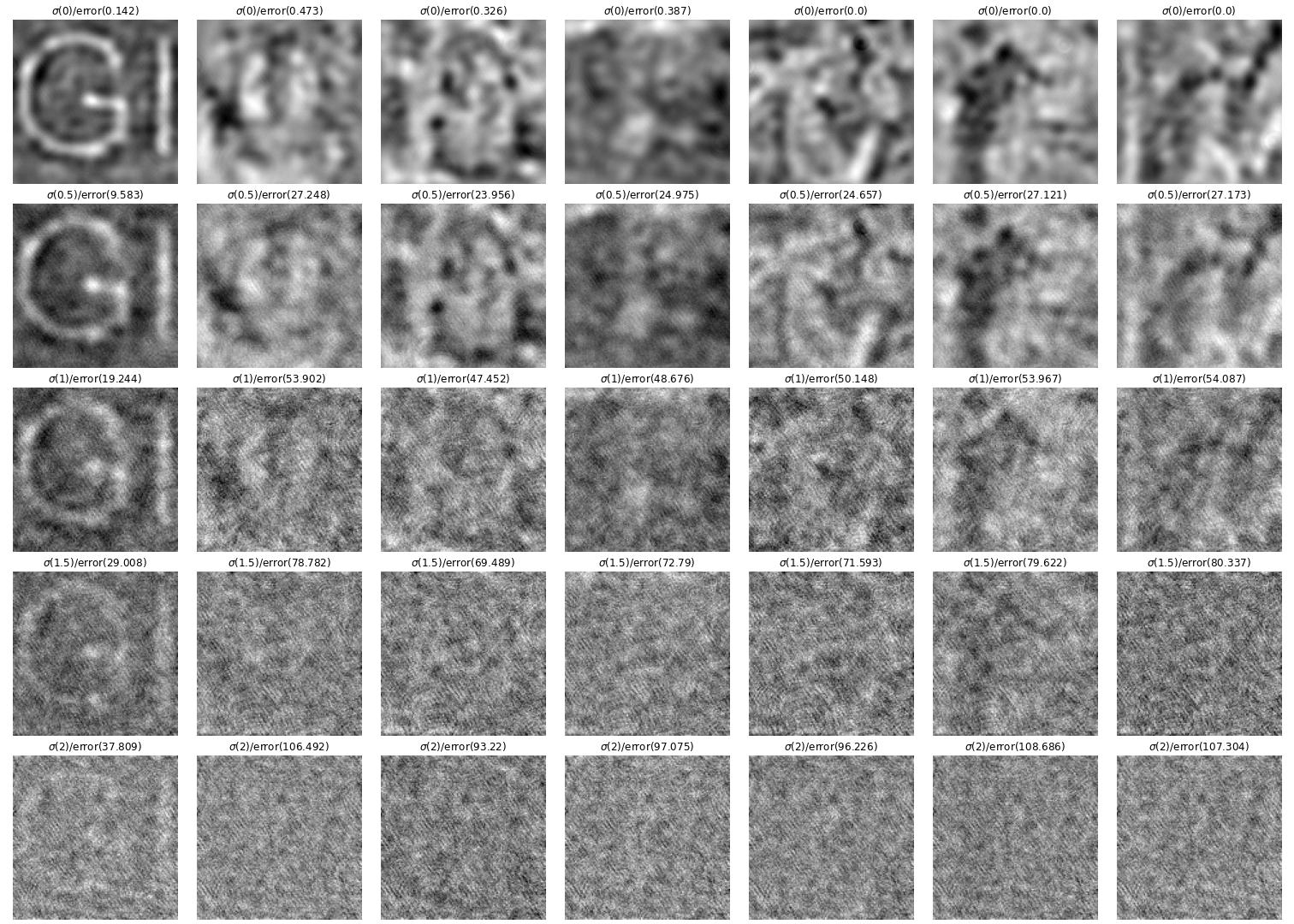}
\caption{Reconstruction results of seven images using constructed $A_{recv}$ by Algo.3 on different gaussian noise levels.}
  \label{fig:calibrate-res}
\end{figure}

To test the calibration \hyperref[algo:calidate-mspace]{Algorithm.\ref{algo:calidate-mspace}}(Algo.3 or Calibration.M), conduct QR decomposition on the transpose of pre-measurement matrix $A^T$, and $M$ column vectors of the $Q\in R^{N\times M}$ matrix form the basis of image space $\mathcal{S}$. In order to ensure that some images in the experiment are always in this $\mathcal{S}$ space, the last three images are used to replace the three bases in the image space $\mathcal{S}$. This operation destroys the orthogonality of the basis. Considering the difficulty of constructing orthogonal basis images in practical applications, this study further explores whether non strict orthogonal basis can still reconstruct the original image. Similarly, add different gaussian noise levels, and the reconstruction results are shown in \hyperref[fig:calibrate-res]{Fig.\ref{fig:calibrate-res}}. It can be seen that under the low noise level of non strict orthogonal basis, the constructed $A_{recv}$ can still reconstruct the original images, although it is not clear enough. This indicates that the calibration algorithm has a certain tolerance for non strict orthogonal basis and noise, but overall, the anti-noise ability of constructed $A_{recv}$ by such algorithm is also weak.

\subsection{Impact of computational Precision}
To understand the impact of computational precision, need to get the core step of \hyperref[algo:1]{Algo.\ref{algo:1}}, \hyperref[algo:2]{Algo.\ref{algo:2}} and \hyperref[algo:calidate-mspace]{Algo.\ref{algo:calidate-mspace}}. Regardless of which algorithm is used, the final constructed $A_{recv}$ is the result of $A_{recv}^{e_y^k}$ accumulation.
\begin{equation}
\label{eq:easum}
A_{recv} = \sum_{k} A_{recv}^{e_y^k}
\end{equation}

Consider use constructed $A_{recv}$ to simulate measure image $x$, obtaining the expression as shown in \hyperref[eq:eamr]{Eq.\ref{eq:eamr}}.

\begin{equation}
\label{eq:eamr}
y \approx y_{recv} = A_{recv}x = \sum_{k} \frac{y_0^T\Sigma Ax}{y_0^T\Sigma y_0}\;e_y^k = \sum_{k} \lambda(y_0,\Sigma,A,x) \;e_y^k
\end{equation}

Since $y_0,\Sigma,A$ are all independent of $x$, they can be treated as constants, leaving only $x$ as the variable. And in matched solution algorithm, $\lambda \in R$ is a constant because of all parameters of $\lambda(y_0,\Sigma,A,x)$ are independent of iteration variable $k$. So extract this constant can yield:

\begin{equation}
y \approx A_{recv}x=\lambda(y_0,\Sigma,A,x)\sum_{k}e_y^k=\lambda(x)\sum_{k}e_y^k
\end{equation}

This means that the solution of reconstruction algorithm $G(y, A_{recv})$ can be any image $x$, which is determined by the multiplier property of constructed $A_{recv}$. If use constructed $A_{recv}$ to simulate measure another arbitrary image $x'$, the measurement value as shown in \hyperref[eq:manxl]{Eq.\ref{eq:manxl}}, which is proportional to the measurement value $y$ obtained by measure $x$. This indicates that $x$ and $x'$ both conform to the measurement equation, which both in the solution space of the reconstruction algorithm $G(y, A_{recv})$ naturally.
\begin{equation}
\label{eq:manxl}
A_{recv}x'=\lambda(x')\sum_{k}e_y^k=\frac{\lambda(x')}{\lambda(x)}\lambda(x)\sum_{k}e_y^k \approx \frac{\lambda(x')}{\lambda(x)}y
\end{equation}

If high-precision matched solution algorithms are used now, the coefficient $\frac{\lambda(x')}{\lambda(x)}$ is a constant scalar, which will result in feasible solution for any image. In this case, most compressed sensing reconstruction algorithms will choose a sparse solution as the optimal solution. On the contrary, if low precision matched solution algorithms are used, the constructed $A_{recv}$ does not strictly satisfy the multiplier property as shown in \hyperref[eq:manxl]{Eq.\ref{eq:manxl}}, which means that \hyperref[eq:manxl2]{Eq.\ref{eq:manxl2}} holds.

\begin{equation}
\label{eq:manxl2}
\begin{aligned}
&y \approx A_{recv}x \\
A_{recv}x' = &\vec{\lambda} \star (A_{recv}x) \approx \vec{\lambda} \star y
\end{aligned}
\end{equation}

$\star$ represents point-wise multiplication of vectors. $\vec{\lambda}$ is a vector with the same dimension of $y$ and it's components have fluctuation, not constant. So at this point, only $x$ satisfies measurement equation, and reconstruction algorithm $G(y, A_{recv})$ can solve for the optimal solution $x$, without confusion from any other image $x'$. And components of $\vec{\lambda}$ have larger fluctuation, the lower any image confusion and the more exact optimal solution $x$.
~\\\\
As for calibration solution algorithm, due to $y_0=Y(:,k)$ is related to iteration variable $k$, $\lambda(y_0,\Sigma,A,x)$ cannot be extracted out of sum. Obviously, its constructed $A_{recv}$ does not satisfy the multiplier property as shown in \hyperref[eq:manxl]{Eq.\ref{eq:manxl}} but rather have \hyperref[eq:manxl2]{Eq.\ref{eq:manxl2}} hold. So, Algo.3 is not impacted by computational precision.
~\\\\
To verify the correctness of above analysis, conduct same experiment on three different devices, which computational precision rank is RTX3090$>$RTXA400$>$ RTX2080 Ti. The same experiment is that \hyperref[algo:1]{Algo.\ref{algo:1}} and \hyperref[algo:2]{Algo.\ref{algo:2}} use GI image as measured images $x$ and other parameters $(PM3, A, A_u)$ remained consistent with previous experiments to construct $A_{recv}$, \hyperref[algo:calidate-mspace]{Algo.\ref{algo:calidate-mspace}} also uses same parameters $(A, A_u)$ to construct $A_{recv}$. Then select Baboon image as other image $x'$ and put pair $(A_{recv},x,x')$ into \hyperref[eq:manxl2]{Eq.\ref{eq:manxl2}} to calculate $\vec{\lambda}$ and its components range.

\begin{figure}[H]
  \centering
  \setlength{\abovecaptionskip}{-0.05cm}
  \includegraphics[scale = 0.42]{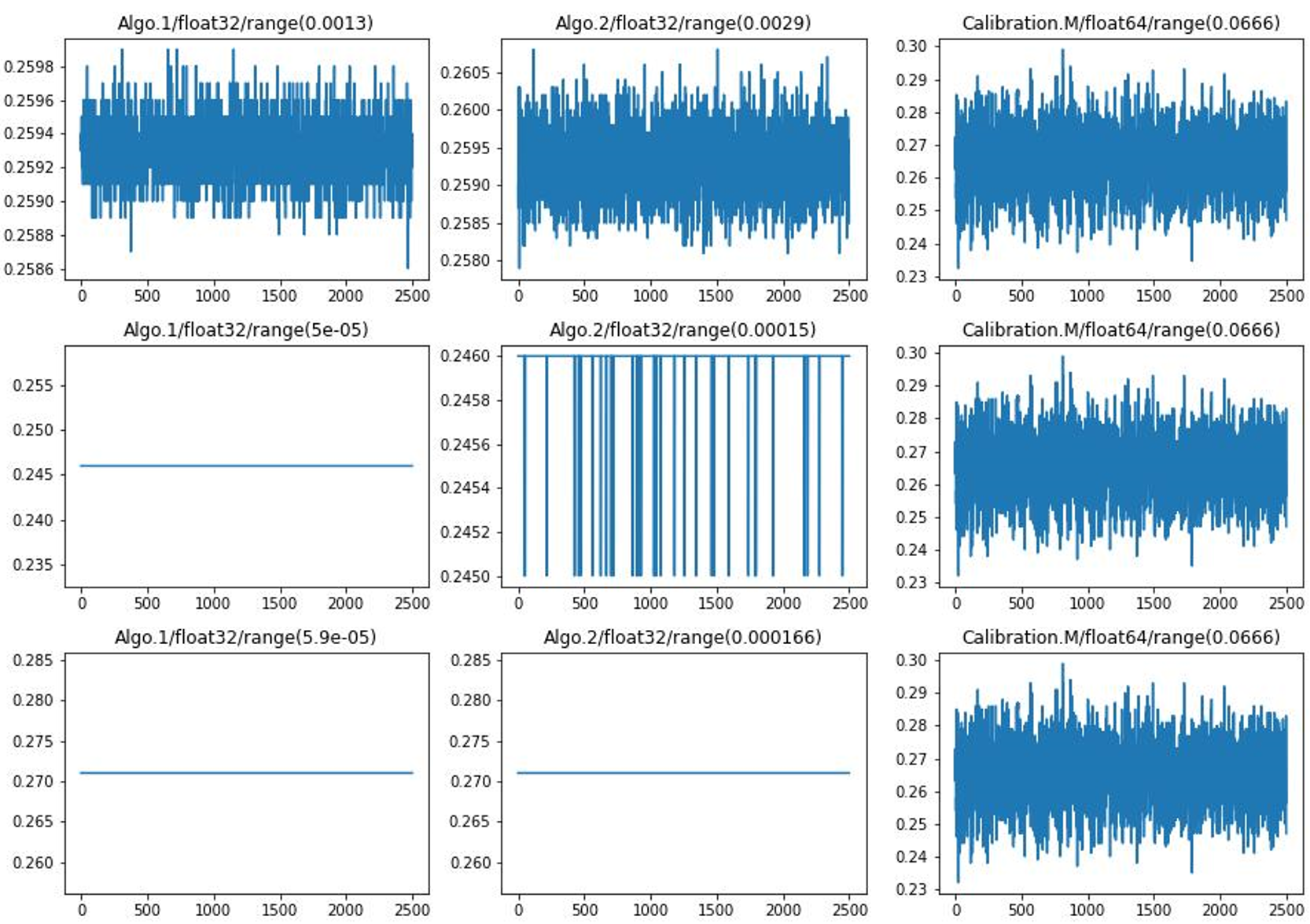}
\caption{$\vec{\lambda}$ components curves of three algorithms on three devices. First line is RTX2080 Ti, second line is RTXA400, third line is RTX3090.}
  \label{fig:mt-curve}
\end{figure}

\begin{figure}[H]
  \centering
  \setlength{\abovecaptionskip}{-0.05cm}
  \includegraphics[scale = 0.7]{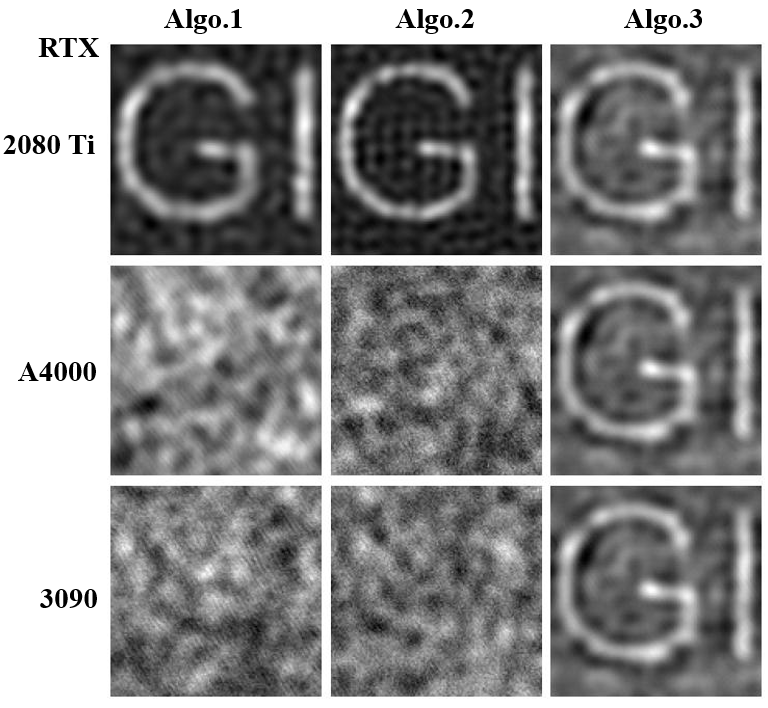}
\caption{Reconstructed GI image using constructed $A_{recv}$ by three algorithms on three devices.}
  \label{fig:mt-res}
\end{figure}

From components curves of \hyperref[fig:mt-curve]{Fig.\ref{fig:mt-curve}} and reconstructed images results of \hyperref[fig:mt-res]{Fig.\ref{fig:mt-res}}, it can be seen that the fluctuation of the $\vec{\lambda}$ component is consistent with the reconstruction. The original image is reconstructed by Algo.1, Algo.2 and Algo.3 on RTX2080Ti, as well as Algo.3 on RTX400 and RTX3090. These components curves fluctuate greatly, and components ranges are also large. The remaining components curves have very small fluctuations or even constants, and so the original images cannot be reconstructed.
~\\\\
Based on above analysis and experiments, it can be concluded that matched solution \hyperref[algo:1]{Algo.\ref{algo:1}} and \hyperref[algo:2]{Algo.\ref{algo:2}} should use float32 for calculations on medium precision devices, while calibration solution \hyperref[algo:calidate-mspace]{Algo.\ref{algo:calidate-mspace}} should use float64 for calculations on higher precision devices as much as possible.

\section{Discussion}
Based on the above experimental results and analysis, three key points can be summarized.

(i) A mismatch reconstruction theory for solving unknown measurement matrix is proposed. The core contribution of this paper lies in proposing the mismatch equation and designing two types of algorithms based on it: matched and calibration solution for unknown measurement matrix. The matched solution uses error iteration algorithm to construct a new measurement matrix $A_{recv}$ for target measurement value, which is highly flexible. The calibration solution constructs a universal $A_{recv}$ that can be applied to any measurement value under the same fiber bending configuration by measuring orthogonal basis images in the image space and combining with the calibration equation. Experimental results have shown that both types of algorithms can successfully reconstruct the original images at low noise levels, verifying the effectiveness of this theoretical framework.

(ii) The reconstruction performance using constructed $A_{recv}$ is affected by noise, computational precision and orthogonality of basis images. Firstly, the reconstruction effect is good in noise free or low noise environments, but the reconstruction quality sharply decreases as the noise level increases. Secondly, the impact of computational precision on matched solution algorithms is particularly critical. On the low precision devices, since constructed $A_{recv}$ does not strictly satisfy multiplier property, it actually helps the reconstruction algorithms find the unique optimal solution. However, on the high precision devices, confusion may be introduced due to the expansion of solution space caused by the multiplier property. For calibration solution algorithm, computational precision has no impact, but its performance is affected by orthogonality of basis images. Although experiment has shown that it has a certain tolerance for non strictly orthogonal basis, the non orthogonality of the basis images can still introduce error, leading to a decrease in reconstruction clarity.

(iii) This theory reduces the stringent requirements for precision calibration equipments and provides a low cost, portable multimode fiber imaging method in dynamic environments. For example, simply using existing imaging measurement systems and algorithms instead of cameras capturing speckles to construct measurement matrix, and quickly conduct matched or calibration algorithms with each disturbance change. This means that in some cost sensitive or harsh working environment applications, such as industrial pipelines and human cavity endoscopic detection equipments, advanced algorithms can be used to compensate for hardware deficiencies and achieve collaborative optimization of software and hardware, which is of great significance for promoting the popularization of advanced imaging technology.

\section{Conclusion}
In this paper, mismatch reconstruction theory is proposed to solve the problem of unknown measurement matrix in imaging through multimode fiber bending. Based on the core of this theory, which is mismatch equation and its multiplier property, matched and calibration solution algorithms are further proposed to construct a new measurement matrix $A_{recv}$ for unknown measurement matrix. The matched solution algorithm uses error iteration method to construct $A_{recv}$ that matches target measurement value under the measurement equation through only one additional measurement. And the calibration solution algorithm constructs $A_{recv}$ by measuring orthogonal basis images in the image space and combining with calibration equation, which is applicable to any measurement value under the same fiber bending configuration. Theoretical analysis and experimental results both show that constructed $A_{recv}$ by these algorithms can be used to reconstruct the original images successfully, and has certain robustness for noise, computational precision and orthogonality. In the future, this theory will be applied to stable dynamic imaging through multimode fiber with arbitrary bending in real environment.

\newpage
\section{References}
\vspace{-2em}
\renewcommand{\refname}{}
\bibliographystyle{unsrt} 
\bibliography{cites}

\begin{thebibliography}{10}

\bibitem{WOS:000235136400007}
G~Cammarota, P~Cesaro, A~Martino, G~Zuccalà, R~Cianci, E~Nista, LM~Larocca,
  FM~Vecchio, A~Gasbarrini, and G~Gasbarrini.
\newblock High accuracy and cost-effectiveness of a biopsy-avoiding endoscopic
  approach in diagnosing coeliac disease.
\newblock {\em ALIMENTARY PHARMACOLOGY \& THERAPEUTICS}, 23(1):61--69, JAN 1
  2006.

\bibitem{WOS:000631182800001}
Zuoming Fu, Ziyi Jin, Chongan Zhang, Zhongyu He, Zhenzhou Zha, Chunyong Hu,
  Tianyuan Gan, Qinglai Yan, Peng Wang, and Xuesong Ye.
\newblock The future of endoscopic navigation: A review of advanced endoscopic
  vision technology.
\newblock {\em IEEE ACCESS}, 9:41144--41167, 2021.

\bibitem{WOS:001511986100004}
Jiali Chen, Shuai Liu, Tianxin Gao, Xiaoying Tang, Hongen Liao, and Yingwei
  Fan.
\newblock Research progress on endoscopic multimodal optical imaging.
\newblock {\em OPTICS AND LASER TECHNOLOGY}, 191, DEC 2025.

\bibitem{WOS:000328078300006}
Michael Hughes, Tou~Pin Chang, and Guang-Zhong Yang.
\newblock Fiber bundle endocytoscopy.
\newblock {\em BIOMEDICAL OPTICS EXPRESS}, 4(12):2781--2794, DEC 1 2013.

\bibitem{WOS:000481541400016}
John~P. Dumas, Muhammad~A. Lodhi, Batoul~A. Taki, Waheed~U. Bajwa, and Mark~C.
  Pierce.
\newblock Computational endoscopy-a framework for improving spatial resolution
  in fiber bundle imaging.
\newblock {\em OPTICS LETTERS}, 44(16):3968--3971, AUG 15 2019.

\bibitem{WOS:000308801100042}
Tomas Cizmar and Kishan Dholakia.
\newblock Exploiting multimode waveguides for pure fibre-based imaging.
\newblock {\em NATURE COMMUNICATIONS}, 3, AUG 2012.

\bibitem{WOS:001021361400002}
Zhong Wen, Zhenyu Dong, Qilin Deng, Chenlei Pang, Clemens~F. Kaminski, Xiaorong
  Xu, Huihui Yan, Liqiang Wang, Songguo Liu, Jianbin Tang, Wei Chen, Xu~Liu,
  and Qing Yang.
\newblock Single multimode fibre for in vivo light-field-encoded endoscopic
  imaging.
\newblock {\em NATURE PHOTONICS}, 17(8):679+, AUG 2023.

\bibitem{oes:oes-2023-0041}
Guangxing Wu, Runze Zhu, Yanqing Lu, Minghui Hong, and Fei Xu.
\newblock Optical scanning endoscope via a single multimode optical fiber.
\newblock {\em Opto-Electronic Science}, 3(3):230041--1--230041--32, 2024.

\bibitem{Palmieri:24}
Luca Palmieri.
\newblock Mode coupling in optical fibers.
\newblock In {\em Optical Fiber Communication Conference (OFC) 2024}, page
  M2A.5. Optica Publishing Group, 2024.

\bibitem{WOS:000466458800006}
Piergiorgio Caramazza, Oisin Moran, Roderick Murray-Smith, and Daniele Faccio.
\newblock Transmission of natural scene images through a multimode fibre.
\newblock {\em NATURE COMMUNICATIONS}, 10, MAY 2 2019.

\bibitem{WOS:000303879700023}
Ioannis~N. Papadopoulos, Salma Farahi, Christophe Moser, and Demetri Psaltis.
\newblock Focusing and scanning light through a multimode optical fiber using
  digital phase conjugation.
\newblock {\em OPTICS EXPRESS}, 20(10):10583--10590, MAY 7 2012.

\bibitem{WOS:000314806600009}
Ioannis~N. Papadopoulos, Salma Farahi, Christophe Moser, and Demetri Psaltis.
\newblock High-resolution, lensless endoscope based on digital scanning through
  a multimode optical fiber.
\newblock {\em BIOMEDICAL OPTICS EXPRESS}, 4(2):260--270, FEB 1 2013.

\bibitem{WOS:000369051200017}
Lyubov~V. Amitonova, Adrien Descloux, Joerg Petschulat, Michael~H. Frosz, Goran
  Ahmed, Fehim Babic, Xin Jiang, Allard~P. Mosk, Philip St.~J. Russell, and
  Pepijn W.~H. Pinkse.
\newblock High-resolution wavefront shaping with a photonic crystal fiber for
  multimode fiber imaging.
\newblock {\em OPTICS LETTERS}, 41(3):497--500, FEB 1 2016.

\bibitem{WOS:000684701400001}
Zaikun Zhang, Depeng Kong, Yi~Geng, Hui Chen, Ruiduo Wang, Zhengshang Da, and
  Zhengquan He.
\newblock Lensless multimode fiber imaging based on wavefront shaping.
\newblock {\em APPLIED PHYSICS EXPRESS}, 14(9), SEP 1 2021.

\bibitem{WOS:001334679400008}
Zhouping Lyu and Lyubov Amitonova, V.
\newblock Wavefront shaping and imaging through a multimode hollow-core fiber.
\newblock {\em OPTICS EXPRESS}, 32(21):37098--37107, OCT 7 2024.

\bibitem{WOS:000319339600123}
Antonio~M. Caravaca-Aguirre, Eyal Niv, Donald~B. Conkey, and Rafael Piestun.
\newblock Real-time resilient focusing through a bending multimode fiber.
\newblock {\em OPTICS EXPRESS}, 21(10):12881--12887, MAY 20 2013.

\bibitem{WOS:000366574400003}
Ruo~Yu Gu, Reza~Nasiri Mahalati, and Joseph~M. Kahn.
\newblock Design of flexible multi-mode fiber endoscope.
\newblock {\em OPTICS EXPRESS}, 23(21):26905--26918, OCT 19 2015.

\bibitem{WOS:000362419900077}
Damien Loterie, Salma Farahi, Ioannis Papadopoulos, Alexandre Goy, Demetri
  Psaltis, and Christophe Moser.
\newblock Digital confocal microscopy through a multimode fiber.
\newblock {\em OPTICS EXPRESS}, 23(18):23845--23858, SEP 7 2015.

\bibitem{WOS:000449972600108}
Alex Turpin, Ivan Vishniakou, and Johannes~D. Seelig.
\newblock Light scattering control in transmission and reflection with neural
  networks.
\newblock {\em OPTICS EXPRESS}, 26(23):30911--30929, NOV 12 2018.

\bibitem{WOS:000446963000002}
Babak Rahmani, Damien Loterie, Georgia Konstantinou, Demetri Psaltis, and
  Christophe Moser.
\newblock Multimode optical fiber transmission with a deep learning network.
\newblock {\em LIGHT-SCIENCE \& APPLICATIONS}, 7, OCT 3 2018.

\bibitem{WOS:000621765000049}
Changyan Zhu, Eng~Aik Chan, You Wang, Weina Peng, Ruixiang Guo, Baile Zhang,
  Cesare Soci, and Yidong Chong.
\newblock Image reconstruction through a multimode fiber with a simple neural
  network architecture.
\newblock {\em SCIENTIFIC REPORTS}, 11(1), JAN 13 2021.

\bibitem{WOS:000749357900016}
Ganesh~M. Balasubramaniam, Netanel Biton, and Shlomi Arnon.
\newblock Imaging through diffuse media using multi-mode vortex beams and deep
  learning.
\newblock {\em SCIENTIFIC REPORTS}, 12(1), JAN 28 2022.

\bibitem{WOS:000275543500006}
S.~M. Popoff, G.~Lerosey, R.~Carminati, M.~Fink, A.~C. Boccara, and S.~Gigan.
\newblock Measuring the transmission matrix in optics: An approach to the study
  and control of light propagation in disordered media.
\newblock {\em PHYSICAL REVIEW LETTERS}, 104(10), MAR 12 2010.

\bibitem{WOS:000310978500004}
Youngwoon Choi, Changhyeong Yoon, Moonseok Kim, Taeseok~Daniel Yang,
  Christopher Fang-Yen, Ramachandra~R. Dasari, Kyoung~Jin Lee, and Wonshik
  Choi.
\newblock Scanner-free and wide-field endoscopic imaging by using a single
  multimode optical fiber.
\newblock {\em PHYSICAL REVIEW LETTERS}, 109(20), NOV 12 2012.

\bibitem{WOS:000423776600018}
Moussa N'Gom, Theodore~B. Norris, Eric Michielssen, and Raj~Rao Nadakuditi.
\newblock Mode control in a multimode fiber through acquiring its transmission
  matrix from a reference-less optical system.
\newblock {\em OPTICS LETTERS}, 43(3):419--422, FEB 1 2018.

\bibitem{WOS:000448939000065}
Lyubov Amitonova, V and Johannes~F. de~Boer.
\newblock Compressive imaging through a multimode fiber.
\newblock {\em OPTICS LETTERS}, 43(21):5427--5430, NOV 1 2018.

\bibitem{WOS:000466160900091}
Mingying Lan, Di~Guan, Li~Gao, Junhui Li, Song Yu, and Guohua Wu.
\newblock Robust compressive multimode fiber imaging against bending with
  enhanced depth of field.
\newblock {\em OPTICS EXPRESS}, 27(9):12957--12962, APR 29 2019.

\bibitem{WOS:000489024000008}
Antonio~M. Caravaca-Aguirre, Sakshi Singh, Simon Labouesse, Michael Baratta, V,
  Rafael Piestun, and Emmanuel Bossy.
\newblock Hybrid photoacoustic-fluorescence microendoscopy through a multimode
  fiber using speckle illumination.
\newblock {\em APL PHOTONICS}, 4(9), SEP 2019.

\bibitem{WOS:000471824700002}
Chaitanya~K. Mididoddi and Michael~R. Hughes.
\newblock Towards high speed needle microscopy through a multimode fiber by
  single pixel imaging.
\newblock In GJ~Tearney, TD~Wang, and MJ~Suter, editors, {\em ENDOSCOPIC
  MICROSCOPY XIV}, volume 10854 of {\em Proceedings of SPIE}. SPIE, 2019.
\newblock Conference on Endoscopic Microscopy XIV, San Francisco, CA, FEB
  02-04, 2019.

\bibitem{WOS:000358737200016}
Martin Ploeschner, Tomas Tyc, and Tomas Cizmar.
\newblock Seeing through chaos in multimode fibres.
\newblock {\em NATURE PHOTONICS}, 9(8):529+, AUG 2015.

\bibitem{WOS:000325547200074}
Salma Farahi, David Ziegler, Ioannis~N. Papadopoulos, Demetri Psaltis, and
  Christophe Moser.
\newblock Dynamic bending compensation while focusing through a multimode
  fiber.
\newblock {\em OPTICS EXPRESS}, 21(19):22504--22514, SEP 23 2013.

\bibitem{WOS:000387537600022}
Sean~C. Warren, Youngchan Kim, James~M. Stone, Claire Mitchell, Jonathan~C.
  Knight, Mark A.~A. Neil, Carl Paterson, Paul M.~W. French, and Chris Dunsby.
\newblock Adaptive multiphoton endomicroscopy through a dynamically deformed
  multicore optical fiber using proximal detection.
\newblock {\em OPTICS EXPRESS}, 24(19):21474--21484, SEP 19 2016.

\bibitem{WOS:000530854700091}
Mingying Lan, Yangyang Xiang, Junhui Li, Li~Gao, Yuanhang Liu, Ziyu Wang, Song
  Yu, Guohua Wu, and Jianxin Ma.
\newblock Averaging speckle patterns to improve the robustness of compressive
  multimode fiber imaging against fiber bend.
\newblock {\em OPTICS EXPRESS}, 28(9):13662--13669, APR 27 2020.

\bibitem{WOS:000384715800062}
Amir Porat, Esben~Ravn Andresen, Herve Rigneault, Dan Oron, Sylvain Gigan, and
  Ori Katz.
\newblock Widefield lensless imaging through a fiber bundle via speckle
  correlations.
\newblock {\em OPTICS EXPRESS}, 24(15):16835--16855, JUL 25 2016.

\bibitem{WOS:000665038900019}
Shuhui Li, Simon A.~R. Horsley, Tomas Tyc, Tomas Cizmar, and David~B. Phillips.
\newblock Memory effect assisted imaging through multimode optical fibres.
\newblock {\em NATURE COMMUNICATIONS}, 12(1), JUN 18 2021.

\bibitem{WOS:000476652500013}
Pengfei Fan, Tianrui Zhao, and Lei Su.
\newblock Deep learning the high variability and randomness inside multimode
  fibers.
\newblock {\em OPTICS EXPRESS}, 27(15):20241--20258, JUL 22 2019.

\bibitem{WOS:000650530200001}
Xuetian Lai, Qiongyao Li, Xiaoyan Wu, Guodong Liu, Ziyang Chen, and Jixiong Pu.
\newblock Mutual transfer learning of reconstructing images through a multimode
  fiber or a scattering medium.
\newblock {\em IEEE ACCESS}, 9:68387--68395, 2021.

\bibitem{WOS:000680277500001}
Shachar Resisi, Sebastien~M. Popoff, and Yaron Bromberg.
\newblock Image transmission through a dynamically perturbed multimode fiber by
  deep learning.
\newblock {\em LASER \& PHOTONICS REVIEWS}, 15(10), OCT 2021.

\bibitem{WOS:001030642400042}
Abdullah Abdulaziz, Simon~Peter Mekhail, Yoann Altmann, Miles~J. Padgett, and
  Stephen McLaughlin.
\newblock Robust real-time imaging through flexible multimode fibers.
\newblock {\em SCIENTIFIC REPORTS}, 13(1), JUL 14 2023.

\bibitem{WOS:000343145200012}
Ori Katz, Pierre Heidmann, Mathias Fink, and Sylvain Gigan.
\newblock Non-invasive single-shot imaging through scattering layers and around
  corners via speckle correlations.
\newblock {\em NATURE PHOTONICS}, 8(10):784--790, OCT 2014.

\bibitem{WOS:000904474500006}
Enlai Guo, Chenyin Zhou, Shuo Zhu, Lianfa Bai, and Jing Han.
\newblock Dynamic imaging through random perturbed fibers via physics-informed
  learning.
\newblock {\em OPTICS AND LASER TECHNOLOGY}, 158(A), FEB 2023.

\bibitem{WOS:000265494900006}
Mario A.~T. Figueiredo, Robert~D. Nowak, and Stephen~J. Wright.
\newblock Gradient projection for sparse reconstruction: Application to
  compressed sensing and other inverse problems.
\newblock {\em IEEE JOURNAL OF SELECTED TOPICS IN SIGNAL PROCESSING},
  1(4):586--597, DEC 2007.

\end{thebibliography}

\newpage
\appendices
\section{Proof of Mismatch Equation}
\label{appe:A}

Considering without noise, unknown and known measurements are as follows:

\begin{subequations}
\label{eq:af}
\begin{align}
\label{eq:a1}
y&=A_u x \tag{a-1} \\
\label{eq:a2}
y_0&=A x \tag{a-2}
\end{align} 
\end{subequations}

Now need to solve a special solution $A_{recv}$ that satisfies $A_{recv}x=y$ from \hyperref[eq:af]{Eq.a}. Assuming exist a matrix $C\in R^{N\times M}$ making $CA$ is invertible. Then multiply both sides of \hyperref[eq:a2]{Eq.\ref{eq:a2}} by $C$.

\begin{equation}
\label{eq:mc}
\begin{aligned}
Cy_0=CA x \\
(CA)^{-1}Cy_0=x 
\end{aligned} \tag{a-3}
\end{equation}

Then substitute the results into \hyperref[eq:a1]{Eq.\ref{eq:a1}}.
\begin{equation}
\label{eq:asenc}
y=A_ux=A_u(CA)^{-1}(Cy_0) \tag{a-4}   
\end{equation}

Assuming exist a matrix $D\in R^{1\times N}$ making $Cy_0D$ is invertible. Then multiply both sides of \hyperref[eq:asenc]{Eq.\ref{eq:asenc}} by $D$.

\begin{equation}
\label{eq:md}
\begin{aligned}
yD=A_u(CA)^{-1}(Cy_0D) \\
yD(Cy_0D)^{-1}(CA)=A_u
\end{aligned} \tag{a-5}
\end{equation}

For the special solution, the invertible condition can be further assumed orthogonal. Then summarize current assumptions, and continue to simplify \hyperref[eq:md]{Eq.\ref{eq:md}} based on these assumptions.

\begin{description}
    \item[1)] $Cy_0D$ is orthogonal
\end{description}

\begin{equation}
\label{eq:a6}
\begin{aligned} 
A_u&=yD(Cy_0D)^{-1}(CA)  \\
&=yD(Cy_0D)^{T}(CA) \\
&=yDD^Ty_0^TC^TCA  \\
&=||D|| yy_0^TC^TCA \nonumber
\end{aligned} \tag{a-6}
\end{equation}

\begin{description}
    \item[2)] $CA$ is orthogonal
\end{description}

Because of $M\ll N$, $A\in R^{M\times N}$ is a row full rank matrix making $AA^T$ is invertible. So $C^TC$ can be solved as follows:

\begin{equation}
\label{eq:a7}
\begin{aligned}
AA^T = AEA^T = A(CA)^T&CAA^T = AA^TC^TCAA^T \\
\Longrightarrow C^TC = & \;(AA^T)^{-1}
\end{aligned} \tag{a-7}
\end{equation}

Substitute the result of \hyperref[eq:a7]{Eq.\ref{eq:a7}} into \hyperref[eq:a6]{Eq.\ref{eq:a6}}.

\begin{equation}
\label{eq:a8}
A_u=||D|| yy_0^T(AA^T)^{-1}A \tag{a-8}
\end{equation}

Substitute the result of \hyperref[eq:a8]{Eq.\ref{eq:a8}} into \hyperref[eq:a1]{Eq.\ref{eq:a1}} to solve coefficients $||D||$.

\begin{equation}
\label{eq:a9}
\begin{aligned}
y=A_ux = ||D|| &yy_0^T(AA^T)^{-1}Ax=||D||yy_0^T(AA^T)^{-1}y_0 \\
&||D|| = \frac{1}{y_0^T(AA^T)^{-1}y_0}
\end{aligned} \tag{a-9}
\end{equation}

So the special solution is:
\begin{gather}
\label{eq:a10}
A_u =\frac{1}{y_0^T(AA^T)^{-1}y_0} yy_0^T(AA^T)^{-1}A \tag{a-10}
\end{gather}

Replace $\Sigma=(AA^T)^{-1} \in R^{M\times M}$ with any non-zero matrix to obtain the general solution.
\begin{gather}
\label{eq:a11}
A_{recv}^{(y_0,y)} =\frac{1}{y_0^T\Sigma y_0} yy_0^T\Sigma A \; (\Sigma\ne 0) \tag{a-11}
\end{gather}

And regardless of the existence of matrices $C$ and $D$ that make orthogonal assumptions 1) and 2) hold, $A_{recv}^{(y_0,y)}$ can be easily verified as follows:

\begin{equation}
\label{eq:pv}
\begin{aligned}
A_{recv}^{(y_0,y)} &= \frac{1}{y_0^T\Sigma y_0}yy_0^T\Sigma A \nonumber \\
A_{recv}^{(y_0,y)}x = \frac{1}{y_0^T\Sigma y_0}yy_0^T&\Sigma A x = \frac{y_0^T\Sigma A x}{y_0^T\Sigma y_0} y = \frac{y_0^T\Sigma y_0}{y_0^T\Sigma y_0} y = y \nonumber
\end{aligned} \tag{a-12}
\end{equation}

\section{Convergence Proof of Error Iteration Algorithm}
\label{appe:B}

Considering proof the convergence of \hyperref[algo:1]{Algorithm.\ref{algo:1}} with noise. Firstly, use mathematical symbols to describe the iterative process of algorithm. Its initial condition is $A_{recv}^0=0, \;\; e_y^0=y, \;\; \epsilon^0=0$, and the $k$th iteration can be expressed as follows:

\begin{subequations}
\begin{align}
\label{eq:b1}
& A_{recv}^{k+1} = A_{recv}^k + A_{recv}^{e_y^k} \tag{b-1}\\
\label{eq:b2}
& e_y^{k}=y-A_{recv}^{k}x-\epsilon^{k} \tag{b-2}
\end{align}
\end{subequations}

Let $y_{recv}^{k} = A_{recv}^{k}x$ and $\lambda^{k} = y-y_{recv}^{k}$. And the constructed matrix for the final output of iteration is:
\begin{equation}
\label{eq:b3}
A_{recv} = \lim\limits_{k\to+\infty} A_{recv}^{k} \tag{b-3}
\end{equation}

So the goal of proof is that the error sequence $\lambda^{k}$ is convergent and the convergence value is 0.
\begin{equation}
\label{eq:b4}
\begin{aligned}
0&=y-A_{recv}x \\
&=y-\lim\limits_{k\to+\infty}{A_{recv}^{k}}x \\
&=\lim\limits_{k\to+\infty}{(y-A_{recv}^{k}x)}    \\
&=\lim\limits_{k\to+\infty}{(y-y_{recv}^{k})} \\
&=\lim\limits_{k\to+\infty}{\lambda^{k}} \\
\end{aligned} \tag{b-4}
\end{equation}

For the convenience of subsequent derivation, conduct some deformation processing on the pre-measure.
\begin{equation}
\label{eq:b5}
\begin{aligned}
y_0=A*PM_{image} = &\;Ax+A(PM_{image} - x) =Ax+\varepsilon \\ 
\Longrightarrow \varepsilon = &\;A(PM_{image} - x)
\end{aligned} \tag{b-5}
\end{equation}

In order to derive the recurrence formula of $\lambda^{k}$, it can be seen from \hyperref[eq:b1]{Eq.\ref{eq:b1}} that the following equation holds:

\begin{equation}
\label{eq:b6}
\begin{aligned}
y_{recv}^{k+1} = A_{recv}^{k+1}x &= A_{recv}^kx + A_{recv}^{e_y^k}x \\
&=A_{recv}^kx + \frac{1}{y_0^T\Sigma y_0} e_y^k y_0^T\Sigma Ax \\
&=A_{recv}^kx + \frac{1}{y_0^T\Sigma y_0} e_y^k y_0^T\Sigma (y_0-\varepsilon) \\
&=A_{recv}^kx + \frac{1}{y_0^T\Sigma y_0} e_y^k y_0^T\Sigma y_0 -\frac{1}{y_0^T\Sigma y_0} e_y^k y_0^T\Sigma \varepsilon \\
&=A_{recv}^kx + e_y^k -\frac{y_0^T\Sigma \varepsilon}{y_0^T\Sigma y_0} e_y^k \\
&=A_{recv}^kx + (1-k_{\varepsilon})e_y^k  \\
&=y_{recv}^{k} + (1-k_{\varepsilon})e_y^k  \\
\Longrightarrow k_{\varepsilon}&=\frac{y_0^T\Sigma \varepsilon}{y_0^T\Sigma y_0}
\end{aligned} \tag{b-6}
\end{equation}

Using \hyperref[eq:b2]{Eq.\ref{eq:b2}} to eliminate $e_{y}^k$ as follows:

\begin{equation}
\label{eq:b7}
\begin{aligned}
e_y^{k}&=y-A_{recv}^{k}x-\epsilon^{k} \\
&=y-y_{recv}^{k}-\epsilon^{k} \\ 
\Longrightarrow y_{recv}^{k+1} &= y_{recv}^k+(1-k_{\varepsilon})e_y^k \\
&= y_{recv}^k+(1-k_{\varepsilon})(y-y_{recv}^k-\epsilon^{k}) \\
&=y_{recv}^k+y-y_{recv}^k-\epsilon^{k}-k_{\varepsilon} * (y-y_{recv}^k-\epsilon^{k}) \\
y-y_{recv}^{k+1} &=\epsilon^{k}+k_{\varepsilon} * (y-y_{recv}^k-\epsilon^{k}) \\
y-y_{recv}^{k+1} &=k_{\varepsilon}(y-y_{recv}^k)+(1-k_{\varepsilon})\epsilon^{k}
\end{aligned} \tag{b-7}
\end{equation}

Then the recurrence formula of $\lambda^{k}$ is:
\begin{equation}
\label{eq:recformula}
\lambda^{k+1}=k_{\varepsilon}\lambda^{k}+(1-k_{\varepsilon})\epsilon^{k} \tag{b-8}
\end{equation}

It can be seen that this is a geometric sequence with bias noise term, and its general expression is:
\begin{equation}
\label{eq:finequ}
\lambda^k = (k_{\varepsilon})^k\lambda^0 + (1-k_{\varepsilon})\sum_{i=0}^{k-1}(k_{\varepsilon})^{i}\epsilon^{k-i} \tag{b-9}
\end{equation}

The absolute value of convergence factor $k_{\varepsilon}$ is always less than 1 as shown in \hyperref[eq:itercond]{Eq.\ref{eq:itercond}}
\begin{equation}
\label{eq:itercond}
||k_{\varepsilon}||=||\frac{y_0^T\Sigma \varepsilon}{y_0^T\Sigma y_0}||=\frac{||y_0^T\Sigma A(PM_{image}-x)||}{||y_0^T\Sigma y_0||}=||1-\frac{y_0^T\Sigma Ax}{y_0^T\Sigma y_0}||<1 \tag{b-10}
\end{equation}

However, due to the presence of bias noise term, in order to achieve better convergence, it is still necessary to choose appropriate special solution $\Sigma$ and pre-measure $y_0$, making the convergence factor $k_{\varepsilon}$ is as small as possible.
~\\\\
Assuming that the measurement noise each time is gaussian distribution $\epsilon^{k}\sim N(\mu,\sigma^2)$ and \textit{i.i.d.} The convergence value of $\lambda^{k}$ is:
\begin{equation}
\label{eq:b11}
\lim\limits_{k\to+\infty}{\lambda^{k}}\sim N(\mu,\frac{1}{k_{\varepsilon}+1}\sigma^2) \tag{b-11}
\end{equation}

Specifically, when there is no noise, this limit tends to 0. And from \hyperref[eq:finequ]{Eq.\ref{eq:finequ}}, it can be seen that in order to minimize the influence of measurement noise, $k_{\varepsilon}$ must tend 0. This is to say:
\begin{gather*}
When\;\; k_{\varepsilon} \in U(0,\delta_1\rightarrow 0) \;\; and \;\; i>0 \Longrightarrow (1-k_{\varepsilon})(k_{\varepsilon})^i \in U(0,\delta_2\rightarrow 0)
\end{gather*}

\begin{figure}[H]
  \centering
  \setlength{\abovecaptionskip}{-0.05cm}
  \includegraphics[scale = 0.27]{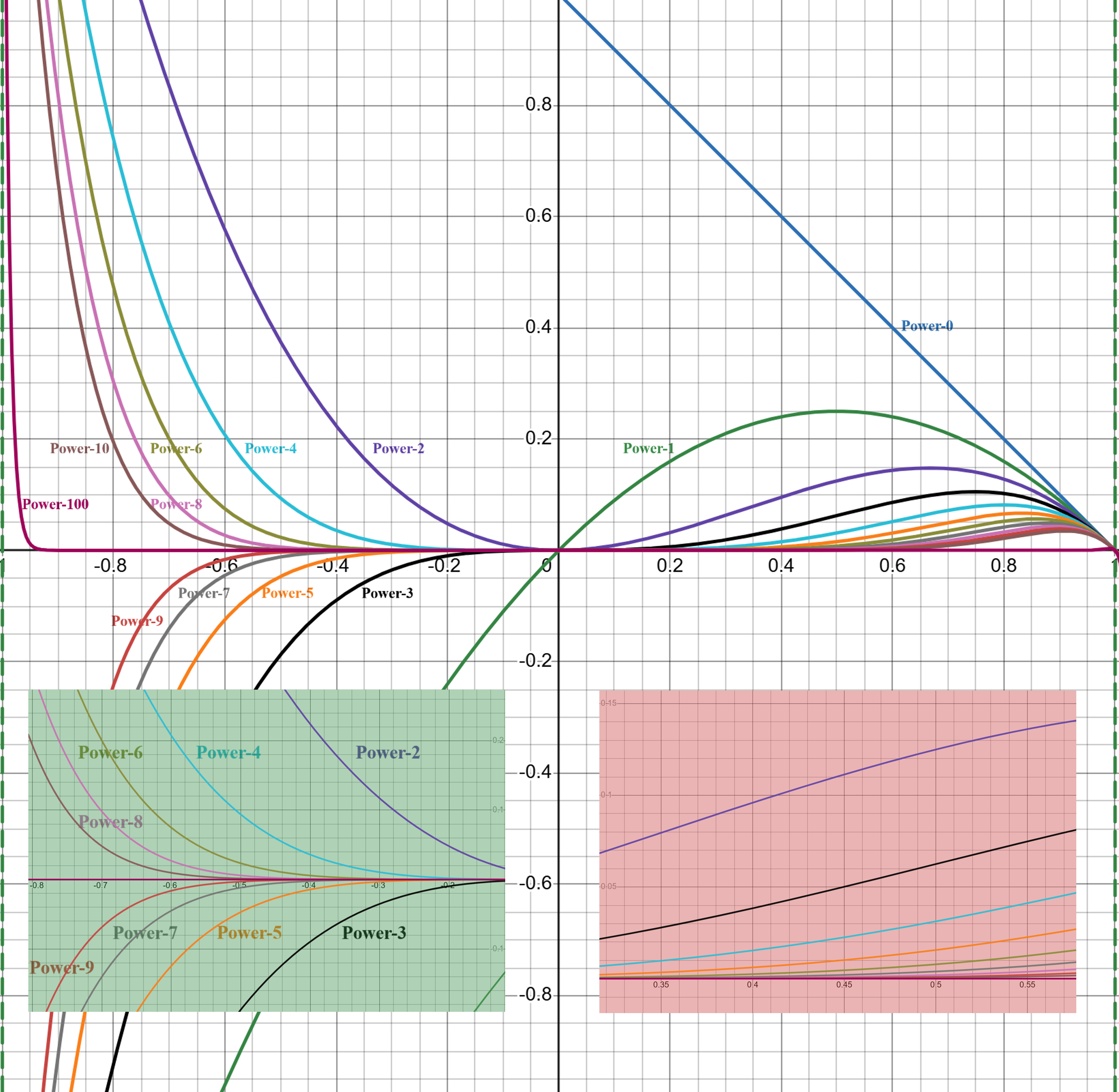}
\caption{Curve family of $\{(1-x)x^{i}\;\vert \;Abs(x)\textless 1\}_{i}$. Left green sub-figure is a scaled figure of parent figure centered at x point of -0.45. Right red sub-figure is a scaled figure of parent figure centered at x point of 0.45.}.
  \label{fig:powers}
\end{figure}

However, the set of function curves $\{(1-x)x^i|Abs(x)<1\}_i$ as shown in \hyperref[fig:powers]{Fig.\ref{fig:powers}}, it can be found that the larger $i$, the wider neighbor radius $\delta_1$. But when $i=0$, noise factor degenerates into $1-k_{\varepsilon}$, which is in $U(1,\delta_2)$. So it impossible to eliminate the last noise $\epsilon^{k}$, and the influence of measurement noise on error iteration algorithm always exists.

\section{Proof of Calibration Equation}
\label{appe:C}

Assuming measured image $x$ can be linearly represented by orthogonal basis images $\{\mathbf{x_i}\}_{i=1}^{D}$.
\begin{equation}
\label{eq:c1}
x=\sum_{i}^{D} b_i\bf{x_i} \tag{c-1}
\end{equation}

Conduct pre-measure and unknown measure on all basis images.
\begin{equation}
\label{eq:c2}
\begin{aligned}
y_{i}^0&=A\mathbf{x_i}\\
y_{i}=&A_u\mathbf{x_i}+\epsilon_{i}
\end{aligned} \tag{c-2}
\end{equation}

The goal is to prove that $A_{recv}$ as shown in \hyperref[eq:c3]{Eq.\ref{eq:c3}} is an exact solution for unknown measurement matrix $A_u$.

\begin{equation}
\label{eq:c3}
A_{recv}=\sum_{j}^{D} a_j A_{recv}^{(y_{j}^0,y_j)} \tag{c-3}
\end{equation}

Calculate the expected measurement value for measured image $x$ under above $A_{recv}$.

\begin{equation}
\label{eq:c4}
y_{recv}=A_{recv}x=\sum_{i}^{D}b_iA_{recv} \mathbf{x_i}=\sum_{i}^{D}b_i\sum_{j}^{D}a_jA_{recv}^{(y_{j}^0,y_j)}\mathbf{x_i} \tag{c-4}
\end{equation}

Further use the multiplier property of mismatch equation to obtain:

\begin{equation}
\label{eq:c5}
\begin{aligned}
A_{recv}^{(y_{j}^0,y_j)}\mathbf{x_i}&=\frac{1}{(y_j^0)^T\Sigma  y_j^0} y_j(y_j^0)^T\Sigma A\mathbf{x_i}=k(i,j)y_j\\
\Longrightarrow k(i,j)&=\frac{(y_j^0)^T\Sigma A\mathbf{x_i}}{(y_j^0)^T\Sigma  y_j^0}=\frac{(y_j^0)^T\Sigma y_i^0}{(y_j^0)^T\Sigma  y_j^0}
\end{aligned} \tag{c-5}
\end{equation}

Then the measurement value $y_{recv}$ can be simplified as follows:

\begin{equation}
\label{eq:c6}
y_{recv}=\sum_{i}^{D}b_i\sum_{j}^{D} a_jk(i,j)y_j=\sum_{j}^{D}(\sum_{i}^{D}b_i*a_j*k(i,j))y_j \tag{c-6}
\end{equation}

On the other hand, the unknown measurement value is:
\begin{equation}
\label{eq:c7}
\begin{aligned}
y=A_ux+\epsilon&=\sum_{i}^{D}b_i A_u\mathbf{x_i}+\epsilon \\
&=\sum_{i}^{D}b_i (y_i-\epsilon_{i})+\epsilon \\
&=\sum_{j}^{D}b_j y_j -\sum_{j}^{D}b_j \epsilon_{j} + \epsilon 
\end{aligned} \tag{c-7}
\end{equation}

In order to make the difference between $y$ and $y_{recv}$ is within the error range, just need to take the appropriate $\{a_j\}_{j=1}^{D}$ and $\Sigma$ satisfy the following condition:

\begin{equation}
\label{eq:ccon}
\forall j \;\;\sum_{i}^{D}b_i*a_j*k(i,j)=b_j \tag{c-8} 
\end{equation}

So the error of measurement value is:
\begin{equation}
\label{eq:y-error}
y_{recv}-y=
\begin{cases}
\epsilon' = \sum_{j}^{D}b_j \epsilon_{j} + \epsilon & nosie\\
0& without \;\; nosie
\end{cases} \tag{c-9}
\end{equation}

The only possible situation to satisfy the condition \hyperref[eq:ccon]{Eq.\ref{eq:ccon}} is:
\begin{equation}
\label{eq:c10}
a_j*k(i,j)=
\begin{cases}
1& \text{i=j}\\
0& \text{i$\neq$j}
\end{cases} \tag{c-10}
\end{equation}

If assume $\forall i\neq j\;\; k(i,j) \neq 0$. Since $k(i,i) \equiv 1$, sequence $\{a_j\}_{j=1}^{D}$ must satisfy the following condition:
\begin{equation}
\label{eq:c11}
a_j=\begin{cases}
1& \text{i=j}\\
0& \text{i$\neq$j}
\end{cases} \tag{c-11}
\end{equation}

Obviously this condition is untenable because of $a_j$ is not related to $i$, and it cannot be equal to two values at the same time. So there is one situation for the choice of $\{a_j\}_{j=1}^{D}$ and $\Sigma$.
\begin{equation}
\label{eq:c12}
\begin{aligned}
\forall j\;\;a_j&\equiv 1 \\
k(i,j)=\frac{(y_j^0)^T\Sigma y_i^0}{(y_j^0)^T\Sigma  y_j^0}&=
\begin{cases}
1& \text{i=j}\\
0& \text{i$\neq$j}
\end{cases} 
\end{aligned} \tag{c-12}
\end{equation}

Equivalent to
\begin{equation}
\label{eq:c13}
(y_j^0)^T\Sigma y_i^0=
\begin{cases}
c\neq0& \text{i=j}\\
0& \text{i$\neq$j}
\end{cases} \tag{c-13}
\end{equation}

Arrange the pre-measurement value $\{(y_j^0)^T\in R^{1 \times M}\}_{j=1}^{D}$ of all basis images by rows to form a matrix $Y\in R^{D \times M}$. And then obtain a more concise representation for the choice of special solution $\Sigma$ as shown in \hyperref[eq:c14]{Eq.\ref{eq:c14}}, where $E$ is the identity matrix.
\begin{equation}
\label{eq:c14}
Y\Sigma Y^{T}=E
\tag{c-14}
\end{equation}

\section{Additional Supplementary}
\label{appe:D}

\end{document}